\documentclass[10pt,journal,compsoc,hidelinks]{IEEEtran}

\usepackage{graphicx}
\usepackage{hyperref}
\usepackage{mathtools}
\usepackage{tikz-cd} 
\usepackage{amsthm}
\usepackage{pgfplots}
\usetikzlibrary{patterns}
\usetikzlibrary{plotmarks}
\pgfplotsset{compat=1.11,
    /pgfplots/ybar legend/.style={
        /pgfplots/legend image code/.code={%
            \draw[##1,/tikz/.cd,bar width=3pt,yshift=-0.2em,bar shift=0pt]
            plot coordinates {(0cm,0.8em)};
        },
    },
}

\usepackage{microtype}
\usepackage{support-caption}
\usepackage{subcaption}
\usepackage[inline]{enumitem}
\usepackage{booktabs}
\usepackage{arydshln}
\usepackage{multirow}
\usepackage{comment}

\usepackage{textcomp}  
\usepackage{scalerel}  

\usepackage[utf8]{inputenc}
\usepackage[T1]{fontenc}

\def\thumsup{\scalerel*{\includegraphics{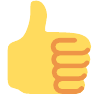}}{\textrm{\textbigcircle}}}

\newcommand{\squishlist}{
 \begin{list}{$\bullet$}
  { \setlength{\itemsep}{0pt}
     \setlength{\parsep}{1pt}
     \setlength{\topsep}{1pt}
     \setlength{\partopsep}{0pt}
     \setlength{\leftmargin}{1.5em}
     \setlength{\labelwidth}{1em}
     \setlength{\labelsep}{0.5em} } }
 \newcommand{\squishend}{\end{list}}

\newcommand{\triple}[1]{$\langle$#1$\rangle$}

\definecolor{darkblue}{rgb}{0.0, 0.0, 0.5}

\newcommand{\newmaterial}{\color{black}}
\newcommand{\oldmaterial}{\color{black}}

\newcommand{\addressreview}{\color{black}}

\renewcommand{\paragraph}[1]{\smallskip\noindent\textbf{#1.\mbox{\ \ }}}

\newcommand{\websiteurl}{https://ascentpp.mpi-inf.mpg.de/}

\newcommand{\bigtableline}{\cmidrule{1-1} \cmidrule(l){2-2} \cmidrule(l){3-3} \cmidrule(l){4-6} \cmidrule(l){7-9} \cmidrule(l){10-10}}

\ifCLASSOPTIONcompsoc
  \usepackage[nocompress]{cite}
\else
  \usepackage{cite}
\fi

\newcommand\ascent{\textsc{Ascent}}
\newcommand\ascentpp{\textsc{Ascent++}}

\begin{document}

\title{Refined Commonsense Knowledge from Large-Scale Web Contents}

\author{Tuan-Phong Nguyen,
        Simon Razniewski,
        Julien Romero, Gerhard Weikum
\IEEEcompsocitemizethanks{\IEEEcompsocthanksitem Tuan-Phong Nguyen, Simon Razniewski and Gerhard Weikum are with the Max Planck Institute for Informatics, Germany. Julien Romero is with Telecom SudParis, France.\protect\\
E-mail: \{tuanphong, srazniew, weikum\}@mpi-inf.mpg.de,\\ julien.romero@telecom-sudparis.eu
}
\thanks{This work has been submitted to the IEEE for possible publication. Copyright may be transferred without notice, after which this version may no longer be accessible.} 
}

\IEEEtitleabstractindextext{%
\begin{abstract}
\newmaterial
Commonsense knowledge (CSK) about concepts and their properties is helpful for AI applications. 
\addressreview
Prior works, 
such as
ConceptNet,
have compiled large CSK collections. 
However, they are restricted in their expressiveness to subject-predicate-object (SPO) triples with simple concepts for S and strings for P and O. \oldmaterial
%
This paper presents a method called \ascentpp{} to automatically build a large-scale knowledge base (KB) of CSK assertions, with refined expressiveness and both better precision and recall than prior works. 
\ascentpp{} goes beyond SPO triples by capturing composite concepts with subgroups and aspects, and by refining assertions with semantic facets. The latter is essential to express the temporal and spatial validity of assertions and further qualifiers.
Furthermore, \ascentpp{} combines open information extraction (OpenIE) with judicious cleaning and ranking by typicality and saliency scores.
For high coverage, our method taps into the large-scale crawl C4 with broad web contents.
The evaluation with human judgments shows the superior quality of the \ascentpp{} KB, and an extrinsic evaluation for QA-support tasks underlines the benefits of \ascentpp{}. A web interface, data, and code can be accessed at 
\textit{\url{\websiteurl}}.
\oldmaterial
\end{abstract}

}

\maketitle

\IEEEdisplaynontitleabstractindextext
\IEEEpeerreviewmaketitle


\IEEEraisesectionheading{\section{Introduction}\label{sec:introduction}}

\paragraph{Motivation} 
Commonsense knowledge (CSK) is a long-standing goal of AI~\cite{mccarthy1960programs,feigenbaum1984knowledge,lenat1995cyc,OpenCyc}: equip machines with structured knowledge about everyday concepts and their properties (e.g., elephants are big and eat plants, buses carry passengers and drive on roads) and about typical human behavior and emotions (e.g., children love visiting zoos, children enter buses to go to school).
%
In recent years, research on the automatic acquisition of CSK assertions has significantly been advanced, and several commonsense knowledge bases (CSKBs) of considerable size have been constructed (see, e.g., \cite{conceptnet,webchild,mishra2017domain,quasimodo}). Use cases for CSK include particularly language-centric tasks such as question answering, conversational systems, and
\addressreview
text generation (see, e.g.,~\cite{Panton2006CommonSR,lin2017reasoning,lin2019kagnet,xia2019incorporating,lin-etal-2020-commongen}). 
\oldmaterial

%
\textit{Examples}: 
Question-answering systems often need CSK as background knowledge for robust answers. For example, when a child asks ``Which zoos have habitats for T-Rex dinosaurs?'', the system should point out that 1) dinosaurs are extinct, and 2) they can be seen in museums, not in zoos.
Dialogue systems should not only generate plausible utterances from a language model but should also be situative, understand metaphors and implicit contexts and avoid blunders. For example, when a user says ``tigers will soon join the dinosaurs'', the machine should understand that this refers to an endangered species rather than alive tigers invading museums.

%

This paper aims to advance the automatic acquisition of CSK assertions from online content to bring better expressiveness, higher precision, and broader coverage.




\paragraph{State-of-the-art and its limitations} 
Large KBs like DBpedia, Wikidata, or Yago focus on encyclopedic knowledge of individual entities like people or places and are sparse on general concepts~\cite{ilievski2020commonsense}. 
Notable projects that concentrate on CSK include ConceptNet~\cite{conceptnet}, WebChild~\cite{webchild}, Mosaic TupleKB~\cite{mishra2017domain}, Quasimodo~\cite{quasimodo},
ATOMIC~\cite{atomic}, TransOMCS~\cite{DBLP:conf/ijcai/ZhangKSR20} and AutoTOMIC~\cite{west2021symbolic}.
%
%
They are all based on 
\addressreview
subject-predicate-object (SPO)
\oldmaterial
triples as knowledge representation and have significant shortcomings:
\squishlist
    \item \addressreview \emph{Expressiveness for subject (S):} \oldmaterial
As subjects, prior CSKBs strongly focus on simple concepts expressed by single nouns (e.g., ``bus'', ``elephant'', ``car'', ``trunk''). This misses semantic refinements 
(e.g., ``diesel bus'' vs. ``electric bus'') 
\oldmaterial
that lead to different properties (e.g., ``polluting'' vs. ``green'') and is also prone to word-sense disambiguation problems (e.g., ``elephant trunk'' vs. ``car trunk''). Even when CSK acquisition considers multi-word phrases, it still lacks the awareness of semantic relations among concepts. Hypernymy lexicons like WordNet or Wiktionary are also very sparse on multi-word concepts. With these limitations, word-sense disambiguation does not work robustly; prior attempts showed mixed results at best (e.g., WebChild~\cite{webchild}, TupleKB~\cite{mishra2017domain}).
    \item \addressreview \emph{Expressiveness for predicate (P) and object (O):} \oldmaterial Predicates and objects are treated as monolithic strings, such as:
    \squishlist
        \item[o] A1: \triple{bus, is used for, transporting people};
        \item[o] A2: \triple{bus, is used for, bringing children to school};
        \item[o] A3: \triple{bus, carries, passengers};
        \item[o] A4: \triple{bus, drops, visitors at the zoo on the weekend}.
    \squishend
This misses the equivalence of assertions A1 and A3 and cannot capture the semantic relation between A1 and A2, namely, A2 refining A1. Finally, the spatial facets of A2 and A4 are cluttered into unrelated strings, and the temporal facet in A4 is not explicit either. The alternative of restricting 
\addressreview
predicate (P)
\oldmaterial
to a small number of pre-specified 
relations
(e.g., ConceptNet~\cite{conceptnet}, WebChild~\cite{webchild}) and 
\addressreview
object (O)
\oldmaterial
to concise phrases comes at the cost of much lower coverage.
    \item \emph{Quality of CSK assertions:} Some of the major CSKBs have prioritized precision (i.e., the validity of the assertions) but have fairly limited coverage (e.g., ConceptNet~\cite{conceptnet}, TupleKB~\cite{mishra2017domain}). Others have broader coverage but include many noisy if not implausible assertions (e.g., WebChild~\cite{webchild}, Quasimodo~\cite{quasimodo}). Very few have paid attention to the saliency of assertions, i.e., the degree to which statements are common knowledge, as opposed to merely capturing many assertions. However, projects along these lines (e.g., ConceptNet~\cite{conceptnet}, ATOMIC~\cite{atomic}) fall short in coverage.
\squishend

\ascentpp{} aims to overcome these limitations of prior works while retaining their positive characteristics. In particular, we aim to reconcile high precision with wide coverage and saliency. Like TupleKB~\cite{mishra2017domain} and Quasimodo~\cite{quasimodo}, we desire to acquire open assertions (as opposed to pre-specified predicates only) but strive for more expressive representations by refining subjects and capturing semantic facets of assertions.
\newmaterial
Furthermore, we also provide a canonicalized version of the resulting CSKB in the ConceptNet schema, thus enabling direct use in applications relying on that fixed schema.

\newmaterial

\paragraph{Approach}
We present the \ascentpp{} method for acquiring CSK assertions with 
refined
semantics from web contents. \ascentpp{} operates in two phases: 
(i) scalable extraction from a large web corpus
and
(ii) aggregation and consolidation.
In the first phase, 
\ascentpp{} 
processes
the C4 crawl~\cite{t5},
a collection of 
365
million
English
web pages.
\ascentpp{} 
extracts 
OpenIE-style tuples by carefully designed dependency-parse-based rules, taking into account assertions for subgroups and aspects of target subjects. 
The extractor
uses
cues from prepositional phrases and adverbs to detect semantic facets and uses
supervised classification for eight facet types. 

In the second phase, on a per-subject basis, \ascentpp{} identifies relevant web pages based on embedding similarity to reference Wikipedia articles, this way being able to distinguish 
homonyms like 
``bus (public transport)'' vs.\ ``bus (network topology)''. \oldmaterial
Assertions are iteratively grouped and 
organized using embedding-based similarity. 
OpenIE-style assertions are canonicalized into the established ConceptNet schema. Finally, a supervised machine learning model ranks the resulting statements by saliency and typicality scores.

We ran \ascentpp{} on the C4 crawl for 10,000 
salient
concepts from ConceptNet as target subjects.
To evaluate the intrinsic quality of the resulting CSKB, we obtained human judgments for a large sample. 
Our
CSKB significantly 
improves over
automatically-built state-of-the-art CSK collections in terms of precision and relative recall.

In addition, we performed an extrinsic evaluation in which commonsense knowledge was used to support language models in question answering. Using three different settings and six different CSKBs, \ascentpp{} significantly outperformed language models without this CSK background knowledge in 2 of the 3 settings, and was best or second best among all 6 CSKBs in all three cases.

\paragraph{Contributions}
This work's key contributions are:
\begin{enumerate}
    \item An expressive model for commonsense knowledge with advanced semantics, subgroups of subjects and faceted assertions as first-class citizens, and scores for typicality and saliency \addressreview (see Section~\ref{sec:knowledge-model}).\oldmaterial
    \item An unsupervised automated method for populating the model with high-quality CSK assertions by large-scale web content extraction and various techniques for aggregation and cleaning \addressreview (see Section~\ref{sec:methodology}).\oldmaterial
    \item Constructing and publicly releasing a high-quality CSKB with 
    2
    million assertions for 10,000 important concepts \addressreview (see Section~\ref{sec:implementation}).\oldmaterial
\end{enumerate}
A web interface to the \ascentpp{} KB, with downloadable data and code, is accessible at \textit{\url{\websiteurl}}.

\paragraph{Prior publication}
This paper substantially extends an earlier conference paper~\cite{ascent}. Major extensions are:
\begin{enumerate}
%
\item The extraction method is completely re-worked. The earlier version operated solely on
top-ranked query results from search engine APIs, whereas \ascentpp{} processes a massive web crawl. This involves new techniques for scalability and quality control in the presence of very noisy web contents
(see Sections \ref{subsec:extraction}, \ref{subsec:filtering}, and \ref{subsec:cleaning}).
%
%
\item We extended the data model to include a new dimension of typicality scores, computed by a supervised regression model that considers assertion facets (see Section~\ref{subsec:ranking}).
%
%
\item We improved our techniques for canonicalizing assertions, added automatic mapping onto the schema of ConceptNet, and updated the evaluation.
This includes a new module for statement canonicalization, enhanced clustering by leveraging SentenceTransformers \cite{sbert}, and evaluation with GPT-3 \cite{GPT3}
(see Sections \ref{subsec:clustering}, \ref{subsec:conceptnet_mapping}, and \ref{subsec:extrinsic}).
\addressreview
\item
Due to this enhanced methodology, the \ascentpp{} KB improves over the previous \ascent{} KB~\cite{ascent} in KB scope and quality. \ascentpp{} has 25\% higher typicality, and 14.6\% higher saliency (for top-10 assertions per subject) and 36.6\% higher relative recall (see Subsection~\ref{sec:ascentpp-vs-ascent}).
\oldmaterial
\end{enumerate}

\oldmaterial

\section{Related Work}

\paragraph{Commonsense knowledge bases (CSKBs)}
CSK acquisition has a long tradition in AI
(e.g., \cite{lenat1995cyc,singh2002open,liu2004conceptnet,DBLP:conf/aaai/GordonDS10}). 

\addressreview

The seminal Cyc project
was the first to construct a large KB that captures commonsense knowledge,
based on hand-crafting assertions by a team of knowledge engineers \cite{lenat1995cyc,Panton2006CommonSR}. OpenCyc~\cite{OpenCyc} was released as a non-commcercial version of Cyc, with much smaller size, though. The last version of this KB, OpenCyc 4.0, was released in 2012. However, this project has been discontinued, and no research license is available anymore. Therefore, we could not include Cyc or OpenCyc in our study.

\oldmaterial

More recently, a number of projects have constructed large-scale collections that are
publicly available, such as ConceptNet~\cite{conceptnet}, WebChild~\cite{webchild}, TupleKB~\cite{mishra2017domain}, Quasimodo~\cite{quasimodo}, ATOMIC~\cite{atomic, comet-atomic-2020}, TransOMCS~\cite{DBLP:conf/ijcai/ZhangKSR20} and AutoTOMIC~\cite{west2021symbolic}.

ConceptNet~\cite{conceptnet} combined CSK collected by human crowdsourcing and knowledge from existing resources such as Cyc, OpenCyc, WordNet, and Wiktionary.
This CSKB contains highly salient information 
for a few pre-specified predicates
(isa/type, part-whole, used for, capable of, location of, plus lexical relations such as synonymy, etymology, derived terms, etc.),
and it is arguably the most widely used CSKB. 
However, it has limited coverage on
many concepts and its ranking of assertions, based on the number of crowdsourcing inputs, is very sparse and unable to discriminate salient properties against atypical or exotic ones
(e.g., listing ``tree'', ``garden'', and ``the bible'' as locations of ``snake'', with similar scores).
ConceptNet does not properly disambiguate concepts, leading to incorrect assertion chains like \triple{elephant, \textit{HasPart}, trunk}; \triple{trunk, \textit{LocationOf}, spare tire}.

WebChild~\cite{webchild}, TupleKB~\cite{mishra2017domain}, and Quasimodo~\cite{quasimodo} devised
fully automated methods for CSKB construction. They use judiciously selected text corpora (incl. book n-grams, image tags, and QA forums) to extract large amounts of SPO triples. 
WebChild builds on hand-crafted extraction patterns, and TupleKB and Quasimodo rely on open information extraction (OpenIE) with subsequent cleaning.
All three are limited to SPO triples.

\newmaterial
ATOMIC~\cite{atomic, comet-atomic-2020} and ASER~\cite{zhang2020aser,aser2} are recent projects on event-centered CSKB construction. While ATOMIC is entirely based on a large-scale human compilation, assertions in ASER were extracted automatically from large text corpora using dependency patterns. Both resources use fixed relations 
(such as \textit{xNeed}, \textit{xWant}, \textit{isAfter}, or \textit{isBefore} in ATOMIC, and \textit{Reason}, \textit{Result}, \textit{Precedence}, or \textit{Succession} in ASER) and are also limited to triples. 
More recently, TransOMCS~\cite{DBLP:conf/ijcai/ZhangKSR20} mined concept-centric triples from ASER. TransOMCS used ConceptNet assertions as seed samples for pattern extraction, hence having the same pre-specified predicates as ConceptNet.

Other notable CSKB projects include CSKG~\cite{ilievski2021cskg} and GenericsKB~\cite{bhakthavatsalam2020genericskb}. CSKG is an attempt to combine seven different CSK resources into an integrated knowledge graph.
GenericsKB, on the other hand, drops attempts at structuring assertions and instead focuses on collecting generic sentences on a per-subject basis.

Our preliminary paper on this work, \ascent~\cite{ascent,ascent-demo},
developed an open information extraction methodology that captures semantic facets and multi-word compounds as subjects, enabling finer-grained knowledge representation and avoiding common disambiguation errors. 
For knowledge source discovery, \ascent{} used commercial search engine APIs. 
This choice ensured precision but reduced recall and strictly limited the scalability on monetary grounds. 
The present  \ascentpp{} approach overcomes this limitation by using 
a huge web crawl
instead,
with several orders of magnitude larger input.


\oldmaterial

\paragraph{Taxonomy and meronymy induction}
The organization of concepts in terms of subclass and part-whole relationships, termed hypernymy and meronymy,
has
received great attention in NLP
and web mining (e.g.,
\cite{DBLP:conf/www/EtzioniCDKPSSWY04,DBLP:conf/acl/SnowJN06,DBLP:journals/coling/GirjuBM06,DBLP:conf/acl/PantelP06,DBLP:conf/acl/PascaD08,DBLP:journals/ai/PonzettoS11,DBLP:conf/sigmod/WuLWZ12,HertlingPaulheim:ISWC2017}).
The hand-crafted WordNet lexicon \cite{wordnet}
organizes over 100k synonym sets for these relationships, although meronymy is sparsely populated.

Recent methods for large-scale taxonomy induction from web sources include 
WebIsADB \cite{seitner2016large,HertlingPaulheim:ISWC2017} building on Hearst patterns and other techniques,
and the industrial GIANT ontology 
\cite{DBLP:conf/sigmod/LiuGNLWWX20}
based on neural learning from user-action logs and other sources.

Meronymy induction at a large scale
has been addressed by \cite{tandon2016commonsense,haspartkb,bhakthavatsalam2020genericskb} 
with pre-specified and automatically learned patterns for refined relations
like physical-part-of, member-of, and
substance-of.

Our approach includes relations of both kinds
by extracting knowledge about
salient subgroups and aspects 
of subjects. 
In contrast to typical taxonomies and part-whole collections, our subgroups include many multi-word phrases: composite noun phrases 
(e.g., ``forest elephant'', ``elephant keeper'') \oldmaterial
and
adjectival and verbal phrases
(e.g., ``male elephant'', ``working elephant''). \oldmaterial
Aspects cover additional refinements of subjects that do not fall under taxonomy or meronymy
(e.g., ``elephant's diet'', ``elephant habitat''). \oldmaterial

\paragraph{Expressive knowledge representation and extraction}
Modalities such as ``always'', ``often'', ``rarely'', and ``never''
have a long tradition in AI research
(e.g., \cite{gabbay2003many}), based on various kinds of modal logics or semantic frame representations, and semantic web formalisms can capture context using, e.g., RDF* or reification~\cite{hoganetal}.
While such expressive knowledge representations have been around for decades, there has hardly been any work that populated KBs with such refined models,
notable exceptions being the 
Knext project
\cite{schubert2002can} at a small scale
and OntoSenticNet \cite{DBLP:journals/expert/DragoniPC18} with a focus on affective valence annotations.

Other projects have pursued different kinds of contextualizations for CSK extraction, notably \cite{zhang2017ordinal}, which scored natural language sentences on an ordinal scale covering the spectrum 
``very  likely'', ``likely'', ``plausible'', ``technically possible'', and ``impossible'',
\cite{chen-etal-2020-uncertain}
with probabilistic scores, and the Dice project
\cite{chalier2020joint} which ranked assertions along 
the dimensions of plausibility, typicality, remarkability, and saliency. 

Semantic role labeling (SRL) is 
a representation and methodology
where sentences are mapped onto
frames (often for certain types of events),
and respective slots 
(e.g., agent, participant, instrument)
are filled with values extracted from the input text
\cite{DBLP:series/synthesis/2010Palmer,clarke2012nlp,semanticrolelabelling}. 
Recently, this paradigm has been extended towards facet-based open information extraction, where extracted tuples
are qualified with semantic facets like location and mode~\cite{stuffie,graphene}.
\ascent{} and \ascentpp{} have built on this general approach but extended it in various ways geared for the case of CSK:
focusing on specifically relevant facets, refining subjects by subgroups and aspects, and aiming to reconcile precision and coverage for concepts as target subjects.



\newmaterial
\paragraph{Pre-trained language models and CSK}
Pre-trained language models (LMs) like BERT~\cite{devlin2019bert} and GPT~\cite{radford2019language, GPT3}
have revolutionized NLP with remarkable advances on many tasks, including those related to commonsense knowledge.
For instance, the LAMA probe~\cite{petroni2019language} aimed to extract commonsense knowledge from masked language models via a cloze-style QA task. COMET~\cite{bosselut2019comet} is an autoregressive language model fine-tuned on existing CSK resources (e.g., ConceptNet and ATOMIC) that is used to predict objects for given subject-predicate pairs, and there are several related works~\cite{cometalt1,cometalt2}.
However, the quality of COMET's generated CSK assertions is often significantly lower than that of its training resources, either manually-curated or web-extracted ones~\cite{nguyen-razniewski-2022-materialized}.

\addressreview
More recently, \cite{west2021symbolic} introduced a knowledge distillation method that extracts CSK from a general language model, GPT-3~\cite{GPT3}, and uses the extracted knowledge to train a smaller commonsense model, COMET~\cite{bosselut2019comet}, which
was shown to perform better than GPT-3 in terms of commonsense capabilities.
The CSK resource extracted from GPT-3, called AutoTOMIC, 
%
and our refined CSKB, \ascentpp{}, have 
fundamental differences.
First, regarding focused domains and sources of knowledge, AutoTOMIC relies on implicit knowledge from GPT-3 and focuses on event-centric knowledge. 
In contrast,
\ascentpp{} extracts concept-centric knowledge explicitly mentioned in a large web corpus.
Second, AutoTOMIC represents knowledge using the traditional SPO triple format, whereas \ascentpp{} uses a refined commonsense knowledge model to capture more expressive assertions.
Third,
as GPT-3 does not provide open access APIs, prompting it to build a large-scale CSKB 
incurs
significant monetary costs \cite{west2021symbolic} compared to extracting knowledge from freely available web crawls.


\oldmaterial

%



\oldmaterial

\section{Knowledge model}
\label{sec:knowledge-model}

Existing CSKBs typically follow a triple-based data model, linking subjects via predicate phrases to object words or phrases. Typical examples from ConceptNet are \triple{bus, \textit{UsedFor}, travel} and \triple{bus, \textit{UsedFor}, not taking the subway}.

Few projects~\cite{webchild,mishra2017domain} have attempted to sharpen such assertions by word sense disambiguation (WSD) \cite{DBLP:journals/csur/Navigli09}, distinguishing, for example, buses on the road from computer buses.
Likewise, only a few projects~\cite{DBLP:conf/aaaifs/GordonS10,zhang2017ordinal,quasimodo,chalier2020joint} have tried to identify salient assertions against correct ones that are unspecific, atypical, or even misleading (e.g., buses used for avoiding the subway or used for enjoying the scenery).
We extend this prevalent paradigm in two significant ways.

\paragraph{Expressive subjects} 
CSK acquisition starts by collecting assertions for target subjects, usually single nouns. This has two handicaps: 1) it conflates different meanings for the same word, and 2) it misses out on refinements and variants of word senses. While some projects~\cite{webchild,mishra2017domain} tried to use word sense disambiguation (WSD) to overcome the first issue, they were inherently limited by the underlying word-sense lexicons (WordNet or Wiktionary). Indeed, they mainly restrict themselves to single nouns. For example, phrases like 
``city bus'', ``tourist bus'',
``newborn elephant'', or ``baby elephant'' 
\oldmaterial
are not present. 

\noindent Our approach to rectify this problem is twofold:
\squishlist
    
    \item \newmaterial First, when discovering source documents for the target subject, we compare the documents with their reference Wikipedia article and only retain documents with better similarity. This way, we can disentangle different senses, for example, of ``bus'' as in ``public transport'' and ``network topology'' themes.
    
    \item 
    Second, when extracting candidates for assertions from the source documents,
    we also capture multi-word phrases as candidates for refined subjects.
    This way, we can acquire {\em isa}-like refinements to create {\em subgroups} of broader subjects (e.g., ``school bus'', ``city bus'', ``tourist bus'', ``circus elephant'', ``elephant cow'', ``domesticated elephant''), and also other kinds of {\em aspects} that are relevant to the general concept (e.g., ``bus' route'', ``bus capacity'', ``elephant tusk'', ``elephant habitat'' or ``elephant keeper''). 
    \oldmaterial
\squishend

Our notion of {\em subgroups} can be thought of as an inverse {\em isa} relation. It goes beyond traditional taxonomies by better coverage of multi-word composites 
(e.g., ``circus elephant'', ``school bus'').  
\oldmaterial
This allows us to improve the representation of specialized assertions such as 
\triple{circus elephants, catch, balls} and \triple{school bus, transports, students}. 
\oldmaterial

Our notion of {\em aspects} includes part-whole relations (PartOf, MemberOf, SubstanceOf)~\cite{DBLP:journals/coling/GirjuBM06,shwartz2018olive,tandon2016commonsense,haspartkb}, but also further aspects that do not fall under the themes of hypernymy or meronymy 
(e.g., ``bus accident'', ``elephant habitat''). \oldmaterial
Note that, unlike single nouns, such compound phrases are rarely ambiguous, so we have crisp concepts without the need for explicit WSD.

\paragraph{Semantic facets} For CSK, assertion validity often depends on specific temporal and spatial circumstances, e.g., 
elephants scare away lions only in Africa or bathe in rivers only during the daytime. 
Furthermore, assertions often become crisper by contextualization in terms of causes/effects and instruments 
(e.g., children ride the bus \dots to go to school, circus elephants catch balls \dots with their trunks). 



To incorporate such information into an expressive model, we contextualize subject-predicate-object triples with semantic {\em facets}. To this end, we build on ideas from research on semantic role labeling (SRL)~\cite{DBLP:series/synthesis/2010Palmer,clarke2012nlp,semanticrolelabelling}.
This line of research has been initially devised to fill hand-crafted frames (e.g., purchase) with values for frame-specific roles (e.g., buyer, goods, price, etc.).
We start with 35 labels proposed in~\cite{stuffie}, a combination of those in the Illinois Curator SRL~\cite{clarke2012nlp}, and 22 hand-crafted ones derived from an analysis of semantic roles of prepositions in Wiktionary.
As many of these are very special, we condense them into eight widely useful roles that are of relevance for CSK: four that qualify the validity of assertions (\textit{degree}, \textit{location}, \textit{temporal}, \textit{other-quality}), and four that capture other dimensions of context (\textit{cause}, \textit{manner}, \textit{purpose}, \textit{transitive objects}).


\addressreview
\paragraph{Quantitative scoring}
Previous works typically quantify the quality of assertions by a single numeric score. ConceptNet, for instance, scores its assertions essentially by the number of annotators that stated them (in most cases, 1). TupleKB employs a supervised model that predicts a [0-1] score capturing statement plausibility.
With \ascentpp{}, we want to empower downstream users to remain flexible in how to rank and use the data. We thus propose a scoring along two dimensions:
\begin{itemize}
    \item \textit{Saliency}: This score captures how spontaneous an assertion comes to the human mind. For example, elephants being used for tourist rides is quite salient, while elephants sleeping at night is less so. Saliency is important to understand which statements matter to humans (e.g., 
    in conversational agents).
    \item \textit{Typicality}: This dimension captures the degree to which an assertion applies to individual instances of a concept, on a per-subject basis. For example, most elephants sleep in most nights, whereas only few elephants give tourists a ride. Typicality is important to understand which utterances make sense (e.g., in question answering).
\end{itemize}
Typicality and saliency are thus orthogonal dimensions, allowing to capture finer properties of commonsense assertions than just frequency or plausibility. Further dimensions could be considered \cite{chalier2020joint}, though we found \textit{plausibility} not to be a dimension of high discriminative utility (implausible statements should rather not even enter CSKBs).

\oldmaterial

These design considerations lead us to the following knowledge model.


\medskip
Let $C_0$ be a set of primary concepts of interest, which could be manually defined or taken from a dictionary. \\
Subjects for assertions include all $s_0 \in C_0$ and judiciously selected
multi-word phrases containing some $s_0$.\\
Subjects are interrelated by {\em subgroup} and {\em aspect} relations: each $s_0$ can be refined by a set of subgroup subjects denoted $sg({s_0})$ and by a set of aspect subjects denoted $asp({s_0})$. The overall set of subjects is $C := C_0 \cup sg_{C_0} \cup asp_{C_0}$.

\vspace*{0.2cm}
\noindent{\bf Definition [Commonsense Assertion]:}\\
A commonsense assertion for $s \in C$ is a sextuple \triple{s, p, o, F, $\pi$, $\theta$} with single-noun or noun-phrase subject $s$, short phrases for predicate $p$ and object $o$, and a set $F$ of semantic facets.
Each facet $(k,v) \in F$ is a key-value pair with one of eight possible keys $k$ and a short phrase as $v$.

Note that a single assertion can have multiple key-value pairs with the same key (e.g., different spatial phrases).

Furthermore, each assertion is accompanied by two [0-1] scores: $\pi$ for saliency and $\theta$ for typicality.
\hspace*{0pt}\hfill $\qed$



\section{Methodology}
\label{sec:methodology}


\begin{figure*}[t]
    \centering
    \includegraphics[width=\textwidth]{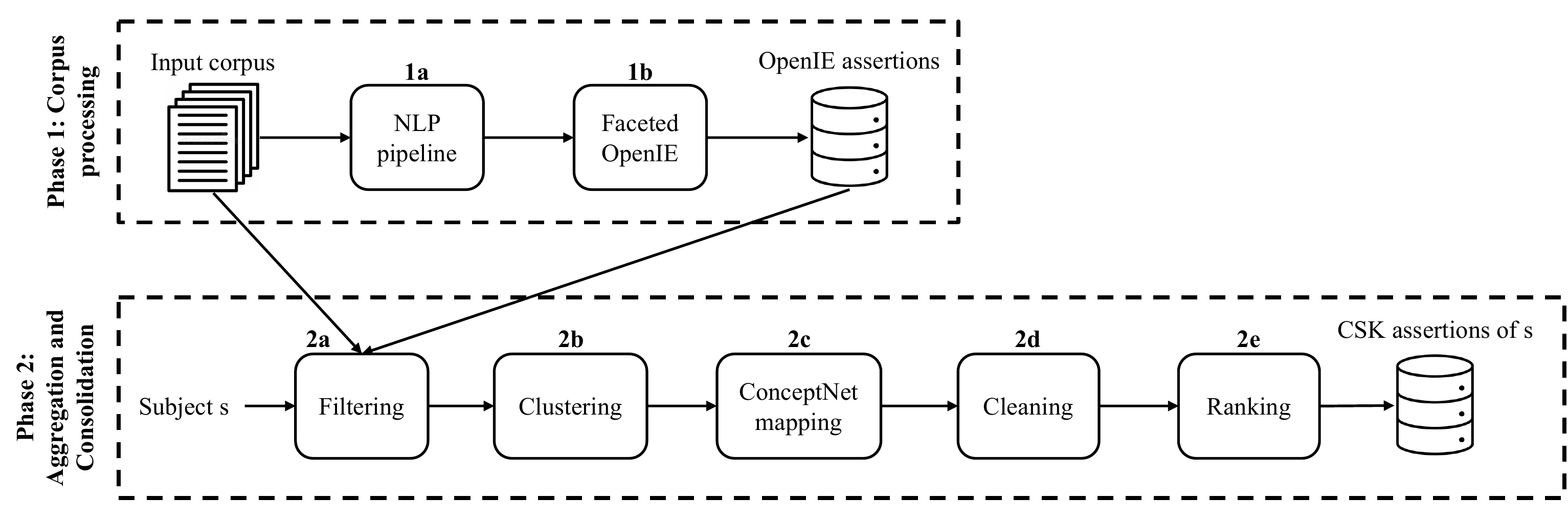}
    \caption{Architecture of the \ascentpp{} system.
    }
    \label{fig:architecture}
\end{figure*}

\subsection{Architecture overview}

\paragraph{Design considerations}
CSK collection has three major design points: (i) the choice of sources, (ii) the choice of the extraction techniques, and (iii) the choice of cleaning or consolidating the extracting candidate assertions.

As \textit{sources}, most prior works carefully selected high-quality input sources, including book n-grams~\cite{webchild}, concept definitions in encyclopedic sources, and school text corpora about science~\cite{clark2018think}. These are often limiting factors in the KB coverage. Moreover, even seemingly clean texts like book n-grams come with surprisingly high noise and bias (cf.~\cite{DBLP:conf/cikm/GordonD13}).
Focused queries for retrieving relevant web pages were used by~\cite{mishra2017domain}, but the query formulations required non-negligible effort. Query auto-completion and question-answering forums were tapped by Quasimodo~\cite{quasimodo, quasimodo-demo}. While this gave access to highly salient assertions, it was, at the same time,
adversely affected by heavily biased and sensational contents 
(e.g., search-engine auto-completion for ``elephants eat'' suggesting ``\dots plastic'' and ``\dots poop''). \oldmaterial

\newmaterial
Our earlier version, \ascent{}~\cite{ascent}, also used search engines combined with refined search queries that use hypernyms for disambiguation. Then, each retrieved web page was compared with a reference document of its corresponding subject to make sure that the web page was not off-topic. This helped \ascent{} collect very high-quality documents for various subjects such as animals (e.g., there are many web pages exclusively talking about animal facts). However, the method did not work well on subjects for which named instances are generally more prominent than the general concepts, such as ``university'' or ``car''. For example, while for ``elephant'', the method extracted  508 assertions with a frequency greater than one, for ``university'', it found a mere 12.
\ascentpp{} rectifies this by directly using a huge corpus as an extraction source. This approach was also taken by ASER~\cite{zhang2020aser} and its derivative, TransOMCS~\cite{DBLP:conf/ijcai/ZhangKSR20}, yet these approaches are recall-oriented and lack appropriate consolidation. 

\oldmaterial
For the \textit{extraction techniques}, choices range from co-occurrence- and pattern-based methods (e.g.,~\cite{elazar2019large}) and open information extraction (e.g.,~\cite{mishra2017domain,quasimodo}) to supervised learning for classification and sequence tagging. Co-occurrence works well for a few pre-specified, clearly distinguished predicates using distant seeds. Supervised extractors require training data for each predicate and thus have the same limitation. Recent approaches, therefore, prefer OpenIE techniques, and our
extractors follow this trend, too.

For \textit{knowledge consolidation}, early approaches kept all assertions from the ingest process (e.g., crowdsourcing~\cite{conceptnet}), whereas recent projects employed supervised classifiers or rankers for cleaning~\cite{mishra2017domain,quasimodo,chalier2020joint,zhang2017ordinal}, and also limited forms of clustering~\cite{mishra2017domain,quasimodo} for canonicalization (taming semantic redundancy).
\newmaterial
In \ascent{}, we used contextual language models to cluster assertions of the same meanings and reinforce the frequency signals of those assertions. 
In the present \ascentpp{}, we apply the same technique with a state-of-the-art model for sentence embeddings, the SentenceTransformers model~\cite{sbert}. Furthermore, after clustering, we do a mapping from our open-schema CSKB to the well-established ConceptNet schema as it is favorable to many researchers (e.g.,~\cite{comet-atomic-2020,lin2019kagnet,feng2020scalable}). Finally, a heuristic-based cleaning approach is applied to eliminate other remaining noise in the resulting KB.

\paragraph{Approach}
The new \ascentpp{} methodology operates in two phases (Fig.~\ref{fig:architecture}):
\squishlist
    \item[1.] \textit{Corpus processing}
    \squishlist
        \item[1a.] \textit{NLP pipeline}: Running NLP pipeline for the input corpus to get NLP features of all sentences, particularly part-of-speech tags and dependency trees.
        \item[1b.] \textit{Faceted OpenIE}: Running the \ascent{} OpenIE system to get faceted assertions from the processed sentences.
    \squishend
    \item[2.] \textit{Aggregation and consolidation}: Each subject is processed separately in this phase.
    \squishlist
        \item[2a.] \textit{Filtering}: Performing a series of document and assertion filtering to get relevant and high-quality assertions for a given subject.
        \item[2b.] \textit{Clustering} of retained assertions based on sentence embeddings.
        \item[2c.] \textit{Mapping} from open assertions into ConceptNet schema.
        \item[2d.] \textit{Cleaning} based on heuristics and a dictionary of unwanted assertions.
        \item[2e.] \textit{Ranking} of assertions: Annotating assertions with complementary scores for typicality and saliency.
    \squishend
\squishend
Both phases treat each document or subject independently and thus can be highly parallelized (see Section~\ref{sec:processing-time}).
\addressreview
Since commonsense knowledge evolves rather slowly (e.g., smartphones emerging over the years), compared to 
encyclopedic knowledge where new entities and relations 
emerge on a daily basis, 
computing a large CSKB is a one-time endeavor with long-term value.
Nonetheless, whenever new or updated inputs need to be processed, Phase 1 can be run incrementally on the new inputs only. 
Only steps 2b and 2e require re-loading previous statements (on a per-subject basis).
Re-running these steps takes about half a day for a corpus like the C4 crawl \cite{t5}.  Table~\ref{tab:statistics-pipeline}
gives detailed run-times (see Section \ref{sec:processing-time}).
\oldmaterial
%

In the following, we discuss Phase 1 and steps 2a-2e of Phase 2 in separate subsections.

\subsection{Corpus processing (Phase 1)
}
\label{subsec:extraction}
Our previous version, \ascent{}, retrieves 500 web pages returned by the Bing Search APIs using hand-crafted search queries for each subject. Usually, not all of those pages would be available for the extraction steps due to crawling errors and another document filtering step in the \ascent{} pipeline. Thus, although this method gathered very high-quality sources, we had to sacrifice recall for several subjects.

We drop it to address these limitations of the web-search component in \ascent{}.
Instead, \ascentpp{} can use any English corpus as its knowledge source. 
This enables the system to tap texts several orders of magnitude larger, making it easy to switch to any particular domain of interest.


\paragraph{Phase 1a: NLP pipeline}
As we reuse the \ascent{} OpenIE approach \cite{ascent}, the NLP pipeline must consist of fundamental operations, including sentence splitting, tokenization, lemmatization, part-of-speech tagging, dependency parsing, and named entity recognition. The extractors will use all of these basic NLP features to output \textit{faceted} OpenIE tuples.

\paragraph{Phase 1b: Faceted OpenIE}
The \ascent{} OpenIE approach was built upon StuffIE~\cite{stuffie}, a rule-based system capable of extracting triples and semantic facets from English sentences. 
The core ideas of the approach are to consider each verb as a candidate predicate of an assertion and then identify subjects, objects, and facets via grammatical relations, so-called dependency paths. 
Subjects must be connected to the candidate predicate through subject-related dependency edges (nsubj, nsubjpass, and csubj) or the adjectival clause edge (acl).
Meanwhile, for objects, the respective edges include direct object (dobj), indirect object (iobj), and nominal modifier (nmod). 
Next, the semantic facets are identified through the following complements to the selected verb: adverb modifiers, prepositional and clausal complements.
We also extended the original set of rules in StuffIE to better deal with conjuncts and adverb facets which helps to identify significantly more assertions and facets and improves the conciseness of the output tuples.
For instance, given the sentence ``elephants use their trunks to pick up objects and drink water'', our system can extract two assertions: \triple{elephants, use, their trunks, \textsc{purpose}: pick up objects} and \triple{elephants, use, their trunks, \textsc{purpose}: drink water}.
\oldmaterial
Details on the dependency-pattern-based rules were previously described in Section 4.2 in \cite{ascent}.


Besides faceted OpenIE tuples, in \ascent{}, we also proposed heuristics to extract fine-grained subjects, i.e., subgroups and aspects, of a given primary concept.

In terms of \textit{subgroups}, they could be sub-species in the case of animals, different types of the primary subject, or refer to the primary subject in different states, 
e.g., ``African elephant'', ``newborn elephant'', ``male elephant'', ``working elephant''.
\oldmaterial
For a primary subject $s_0$, we collect all noun chunks ending with $s_0$ or any of its WordNet lemmas as potential candidates.
Then, we cluster these terms based on word2vec~\cite{mikolov2013distributed} embeddings.
In addition, we leverage WordNet to distinguish antonyms with which the vector space embeddings typically struggle.

For \textit{aspects}, given a primary subject $s_0$, its aspects are extracted from noun chunks collected from two sources:
\begin{enumerate*}[label=(\roman*)]
    \item \textit{possessive noun chunks}, 
    for example, ``elephant's diet'' and ``their diet'' (with a resolution to ``elephant'');
    \item \triple{$s$, $p$, noun chunk} triples where $s$ is equal $s_0$ or any of its WordNet lemmas, and $p$ is one of the following verb phrases: ``have'', ``contain'', ``be assembled of'' or ``be composed of''.
\end{enumerate*}

The extracted subgroups and aspects are also normalized and cleaned in order to avoid spurious extractions or overly specific terms (see Section 4.2 in \cite{ascent}).



\subsection{Filtering (Phase 2a)}\label{subsec:filtering}
While the earlier search-engine-based document retrieval in \ascent{} helped for topical relevance (a core focus of search engines), using a general web corpus as an extraction source implies the presence of substantially more irrelevant content.

Given a subject $s$, one might first collect all OpenIE assertions whose subject is equal $s$ as candidate assertions and then apply ranking or filtering techniques afterward. Similar approaches have been used in Quasimodo~\cite{quasimodo} and TupleKB~\cite{mishra2017domain}. Nevertheless, such post-hoc filtering misses out on broader context from the original documents. In \ascentpp{}, we thus employ some filters first, at the document level, to decide which documents to use for candidate extraction at all. We only extract statements for a primary subject $s_0$ and its subgroups and aspects from a document $d$ if it passes the following filters:
\begin{enumerate}
    \item Document $d$ is only used if it contains between 3 and 40 assertions for $s_0$. The rationale for filters in either direction is that if $s_0$ occurs too rarely, $d$ is more likely off-topic. On the other hand, if $s_0$ occurs too often, then $d$ may be a noisy document such as a machine-created shopping catalog or simply a crawling error.
    \item 
    Then, we compute the cosine similarity between the embeddings of $d$ and the Wikipedia article of the subject $s_0$. Document $d$ will be retained only if the similarity is higher than 0.6 (chosen based on tuning on withheld data).
    This way, we can deal with ambiguous subject terms like ``bus'', which can be either a vehicle or a network topology. \oldmaterial
\end{enumerate}
After the previous filters, we come up with a list of documents for the subject $s_0$. Then, we collect all OpenIE assertions for $s_0$ and its subgroups and aspects from those documents. 
After aggregating the extracted assertions, we only retain those with a frequency of at least 3. That helps to remove noise and redundancy from the results.


\subsection{Clustering (Phase 2b)}\label{subsec:clustering}
Natural language is rich in paraphrases. 
For example, ``elephant eats grass'' can also be written as ``elephant feeds on grass'' or ``elephant consumes grass``.
\oldmaterial
Therefore, identifying and clustering such assertions is necessary to avoid redundancies and get better frequency signals for individual assertions.


For triple clustering, we use the hierarchical agglomerative clustering (HAC) algorithm along with LM-based embeddings to group semantically similar triples.
Specifically, we use
SentenceTransformers~\cite{sbert}, a state-of-the-art architecture for sentence embeddings, to compute embeddings of CSK triples. 
First, given a triple, we concatenate its subject, predicate, and object. Next, we feed the whole string to SentenceTransformers to get its contextualized embeddings. 
Then, for each pair of triples, we compute the Euclidean distance between their normalized embeddings. 
These distances will be used as input for the HAC algorithm.

For facet clustering, we use average word2vec embeddings and the HAC algorithm, the same approach previously used in \ascent{}.
Although more advanced language models such as PhraseBERT~\cite{wang-etal-2021-phrase} could be used instead of word2vec, we found that, for such a limited number of candidate facets (usually less than 10 facets per assertion), word2vec embeddings already provide good 
performance.

\addressreview
The set of chosen hyper-parameters for the clustering algorithms will be presented in Section~\ref{sec:hyperparameters}.

\oldmaterial

\subsection{ConceptNet mapping (Phase 2c)}
\label{subsec:conceptnet_mapping}

\paragraph{Motivation}
There are two main schools for knowledge representation in CSKBs: those relying on open predicates and those using a fixed set. Each has its strengths and challenges regarding expressiveness, redundancy, and usability. To bridge the two, we provide our \ascentpp{} KB in two variants: with open assertions and with canonicalized predicates. Open information extraction supplies the former; the module presented in this subsection
normalizes open assertions into fixed relations.

Our fixed schema of choice is the established ConceptNet schema~\cite{conceptnet}, from which we use the following 19 
relations: \textit{AtLocation}, \textit{CapableOf}, \textit{Causes}, \textit{CreatedBy}, \textit{DefinedAs}, \textit{Desires}, \textit{HasA}, \textit{HasPrerequisite}, \textit{HasProperty}, \textit{HasSubevent}, \textit{IsA}, \textit{MadeOf}, \textit{MotivatedByGoal}, \textit{PartOf}, \textit{ReceivesAction}, \textit{RelatedTo}, \textit{SimilarTo}, \textit{SymbolOf} and \textit{UsedFor}.

Mapping open triples to a fixed schema raises several challenges. In the most straightforward case, the subject and object from the open assertion can remain unchanged, and we only need to pick one of the fixed relations. 
For example, \triple{elephant, lives in, the wild} can be mapped to \triple{elephant, \textit{AtLocation}, the wild}.
\oldmaterial
In some cases, part of the relation and object can be moved, e.g., \triple{elephant, is, a part of a herd} can be mapped to \triple{elephant, \textit{PartOf}, herd}.
\oldmaterial
In other cases, part or all of the predicate is in the object, like in \triple{circus elephant, catches, balls} that can be mapped to \triple{circus elephant, \textit{CapableOf}, catch balls}. 
\oldmaterial

Our normalization method consists of two steps:
\begin{enumerate*}[label=(\roman*)]
    \item first, we use a multi-class classifier to predict a fixed predicate for each open triple;
    \item second, we use a list of crafted rules to modify the object so that the new triple preserves the meaning of the original triple.
\end{enumerate*}

\paragraph{Supervised classifier and rule-based disambiguation}
For the classification model, we fine-tune a RoBERTa model to predict one of the ConceptNet predicates given an input in the following format: \texttt{[CLS] S [SEP] P O}, whereas \texttt{S}, \texttt{P}, and \texttt{O} are the subject, predicate, and object of the open triple. The contextualized embeddings of the \texttt{[CLS]} token will be fed to a fully connected layer to get the prediction. The training data for this model is constructed from ConceptNet triples. Specifically, for each ConceptNet predicate, we 
manually
compiled a list of 1-6 open predicates with similar meanings. For instance, \textit{UsedFor} is aligned with ``be used for'', meanwhile \textit{CapableOf} is aligned with ``be capable of'', ``be able to'', ``can'', ``could'', and an empty string. This way, we can automatically generate approximately 1.2M training examples.

\addressreview
Due to the nature of ConceptNet, the generated data is highly biased towards a few top predicates.
The three most popular predicates are \textit{IsA} (27.92\%), \textit{AtLocation} (20.24\%) and \textit{CapableOf} (13.76\%). 
Meanwhile, the three least popular ones are \textit{MadeOf} (0.20\%), \textit{CreatedBy} (0.06\%) and \textit{SymbolOf} (12 samples, less than 0.001\%).
This imbalance affects the predictions.
The most important difficult case is distinguishing 
the three predicates \textit{IsA}, \textit{HasProperty}, and \textit{ReceivesAction}, which can all be expressed with the open 
predicate word
``be''. 
In the ConceptNet-based training data, the \textit{IsA} relation (27.92\% of the data) is dominant over the other two, \textit{HasProperty} (3.07\%) and \textit{ReceivesAction} (2.20\%).
Therefore, the LM occasionally classifies triples whose predicate is ``be'' incorrectly as \textit{IsA} relations.
\oldmaterial
For this particular case, we have a post-processing step to adjust the predicted predicate: we only assign \textit{HasProperty} to objects which are adjective phrases, \textit{ReceivesAction} to objects which are verb phrases in passive form, and the rest are assigned to \textit{IsA}. The \textit{IsA} relation still contains considerable noise. Hence, later in the cleaning phase (see Section~\ref{subsec:cleaning}), we introduce heuristics to get high-quality assertions of this type.

\addressreview
An alternative could be to re-balance the training data by undersampling the frequent classes or adopting a loss function that gives different weights to different classes. While such generic 
techniques could be considered,
our experience is mirrored in related projects on
creating and curating high-quality KBs,
where injecting a modest amount of expert knowledge is
often the most effective solution
\cite{bhakthavatsalam2020genericskb,yago4}. More advanced methods for relation alignment also exist \cite{soderland2013open,galarraga2014canonicalizing,putri2019aligning,zhang2019openki}.
However, given that ConceptNet itself is imbalanced and has its peculiarities, 
our customized mapping is far superior to
generic alignment methods.
\oldmaterial

\paragraph{Processing objects}
Once we get a predicted fixed-schema predicate, we have to produce an object for the normalized triple. 
The most common case when the object needs modifications is when the predicted predicate is \textit{CapableOf}. 
In this case, if the open predicate is neither \textit{``be capable of''}, \textit{``be able to''}, \textit{``can''} nor \textit{``could''}, the new object will be the concatenation of the original predicate and object. In some cases, a part of the original object must be cut out as it overlaps with the ConceptNet predicate. 
Those predicates include \textit{PartOf} and \textit{SymbolOf}.
Open triples corresponding to those predicates usually look like \triple{elephant, is, part of a herd} or \triple{elephant, is, symbol of strength} which should be 
canonicalized into
\triple{elephant, \textit{PartOf}, herd} and \triple{elephant, \textit{SymbolOf}, strength}, respectively. 
\oldmaterial
Other rules deal with other predicates such as \textit{Desires}, \textit{HasProperty}, \textit{IsA}, \textit{ReceivesAction}, and \textit{UsedFor}.
In total, we have compiled 7 rules for object modification.
We preserve the original object if a triple does not fall into one of these rules.

\subsection{Cleaning (Phase 2d)}\label{subsec:cleaning}

Noise can come from various sources, including OpenIE errors, too specific or general statements, nonsensical assertions, schema normalization errors, etc. Unlike supervised classifiers employed in other projects, our cleaning module is rule-based and thus highly scrutable. \ascentpp{}'s cleaning module consists of the following heuristics:
\begin{enumerate}
    \item First, we compute the perplexity of all triples. To do so, we first turn the triples into sentences and then use an autoregressive model (GPT-2) to compute the perplexity of the sentences. Only triples with medium-to-low perplexity will be retained (in our experiments, we retained triples with perplexity less than 500). 
    \item The \textit{IsA} triples produced by OpenIE are rich but rather noisy. Extracting \textit{IsA} relations is a well-established research theme in entity typing and taxonomy construction and has already reached high-quality results. Therefore, we use the following cleaning algorithm. For each subject, we take the list of its \textit{IsA} relations in ConceptNet
    and extract all head nouns of the objects. For example,
    from the ConceptNet triple \triple{elephant, \textit{IsA}, placental mammal}, we extract the head noun ``mammal''.
    Then, in \ascentpp{}, we only retain the \textit{IsA} assertions in which objects contain one of the trustworthy head nouns extracted earlier. This way, we not only get high-precision assertions but also better recall than ConceptNet.
    If a subject does not occur in ConceptNet, we remove all of its extracted \textit{IsA} assertions, as in our observation, there are usually more noisy \textit{IsA} assertions than valuable ones in the original extraction set.
    Note that, while ConceptNet is our source of choice here, any other existing resources that provide high-quality \textit{IsA} relations can be used, for example, WordNet~\cite{wordnet} or BabelNet~\cite{navigli2012babelnet}.
    \item Then, we manually constructed a dictionary of unwanted objects from our observations. For example, we removed URLs, pronouns, numbers, only stopwords, too general/specific phrases such as ``make sure'' or ``this case'', and vague predicate-object pairs (e.g., \textit{MadeOf}-``part'', \textit{HasProperty}-``available''). We also eliminate ethnicity- and religion-related assertions to avoid potentially critical biases~\cite{mehrabi2021lawyers}.
    \item Finally, we only keep the 1,000 most frequent assertions per subject. For each assertion, we only keep its three most frequent facets. Cutting the tail this way improves the precision significantly but only minorly affects recall (see Subsection~\ref{sec:ascentpp-vs-ascent}).
\end{enumerate}
\addressreview
These filters are a pragmatic technique, and alternatives are conceivable. Some parts require manual work by a knowledge engineer. This holds particularly for the domain-specific dictionary filter. However, the dictionary is small, and a knowledge engineer can easily construct it within a day (at much lower energy consumption and carbon footprint than trying to automate everything computationally).
\oldmaterial

\subsection{Ranking (Phase 2e)}
\label{subsec:ranking}

Existing resources mainly provide unidimensional rankings of their assertions by either \textit{frequency} (e.g., ConceptNet, \ascent) or supervised models trained to predict \textit{plausibility/typicality} (TupleKB, Quasimodo, TransOMCS). However, these two dimensions are quite different, and we consider it important to differentiate their semantics.

\textit{Typicality} states that an assertion holds for most instances of a concept. For example, elephants using their trunk is typical, whereas elephants drinking milk holds only for baby elephants.
\textit{Saliency} refers to the human perspective of whether an assertion is associated with a concept by most humans more or less on first thought.
For example, ``elephants have trunks'' is salient, whereas ``elephants pass by zebras'' is not.


In \ascentpp{}, we make both dimensions first-class citizens of the CSKB and annotate each assertion with scores for both.


\paragraph{Saliency ranking}
For saliency, we rely on the reporting frequency of the assertions, which approximates very well how prominent an assertion is. We transform raw frequencies to a normalized log-scaled frequency as follows:
\begin{equation}\label{eq:saliency}
\small
saliency(spo) = \frac{log(count(spo)) - log(min\_count(s))}{log(max\_count(s))-log(min\_count(s))}
\end{equation}
where $count(spo)$ is the raw frequency of the triple $spo$, $min\_count(s)$ is the minimal frequency of a triple whose subject is $s$, and $max\_count(s)$ is the maximal one.

\paragraph{Typicality ranking}
Typicality is the most challenging dimension. Although reporting frequency correlates with typicality moderately, reporting bias in texts~\cite{DBLP:conf/cikm/GordonD13} means that sensational statements may be grossly overreported and commonalities underrepresented. Our qualitative facets (see Section~\ref{sec:knowledge-model}) give us a unique handle to obtain further insights into typicality.

\noindent We use a linear regression model on three features: 
\begin{itemize}
    \item \textit{Modifier score}. This feature is based on 
    adverbs and quantifiers in facets and subjects. Specifically, we assign every frequency-related 
    modifier
    a specific numeric score (see Table~\ref{tab:modifiers}), then average all scores in each assertion cluster. We consider two types of modifiers: adverbs (e.g., ``always'', ``often'') that appear in semantic facets, and subject quantifiers such as ``all'', ``few'' or ``some''. Without any modifier, we assign a default score of 0.5.
    \item \textit{Neutrality}.
    First, we use a sentiment analysis model\footnote{\url{https://huggingface.co/cardiffnlp/twitter-roberta-base-sentiment}} to compute the probability of a given sentence being positive, neutral, or negative. Then, for each assertion, we consider the average polarity over all of its source sentences as its polarity.
    The value of this feature is $1$ if the assertion is
    classified as neutral.
    Otherwise, a value of zero reflects a polarized assertion.
    \item \textit{Normalized frequency}. The value for this feature is computed as in Equation~\ref{eq:saliency}.
\end{itemize}

\begin{table}[t]
\centering
\begin{tabular}{llr}
\toprule
\textbf{Frequency adverb} & \textbf{Subject quantifier} & \textbf{Score} \\
\midrule
always & all, every & 1.0 \\
\midrule
typically, mostly, mainly & most & 0.9 \\
\midrule
\begin{tabular}[c]{@{}l@{}}usually, normally, regularly, \\ frequently, commonly\end{tabular} & & 0.8 \\
\midrule
 & many & 0.7 \\
 \midrule
often & & 0.6 \\
\midrule
 & some & 0.5 \\
 \midrule
sometimes & & 0.4 \\
\midrule
occasionally & few & 0.3 \\
\midrule
 & & 0.2 \\
 \midrule
hardy, rarely & & 0.1 \\
\midrule
 & no, none & 0.0 \\
 \bottomrule
\end{tabular}
\caption{Frequency modifiers and their scores.}
\label{tab:modifiers}
\end{table}






\newmaterial
\addressreview
\section{Implementation}
\label{sec:implementation}

In this section, we present the implementation of \ascentpp{}, which includes the choice of input corpus, input subjects and other (hyper)-parameters, the processing time of each module, and the size of the resulting CSKB.

\begin{table*}[t]
    \centering
    \begin{tabular}{llcc}
    \toprule
    \textbf{\#} & \textbf{Phase} & \textbf{Processing time} & \textbf{Output} \\
    \midrule
    \textbf{1a} & NLP pipeline & 1.5 days & 365M processed documents \\
    \textbf{1b} & Faceted OpenIE & 20 hours & 8M OpenIE assertions \\
    \textbf{2a} & Filtering & 2 hours & 165M assertions $\rightarrow$ 15M unique triples \\
    \textbf{2b} & Clustering & 10 hours & 7.5M assertion clusters \\
    \textbf{2c} & ConceptNet mapping & 30 minutes & 7.5M canonicalized assertions \\
    \textbf{2d} & Cleaning & 10 minutes & 2M assertions \\
    \textbf{2e} & Ranking & 2 hours & 2M ranked assertions \\
    - & \textbf{Total} & \textbf{\textasciitilde{} 3 days} & \\
    \bottomrule
    \end{tabular}
    \caption{Processing time and output size of each step in the pipeline.}
    \label{tab:statistics-pipeline}
\end{table*}

\subsection{Input text corpus and subjects}
Choosing the input text corpus is essential because large text corpora, especially those scraped from the web, often come with a large portion of noise and irrelevant contents.
Our input corpus of choice is the Colossal Clean Crawled Corpus (i.e., the C4 corpus), created to train the T5 text-to-text Transformers model~\cite{t5}. C4 was created by intensively cleaning the Common Crawl's web crawl corpus. The filtering process includes deduplication, English-language text detection, removing pages containing source code, offensive language, or too few contents and lines, removing lines that did not end with a terminal punctuation mark, etc.
That resulted in a large corpus of 750GB of text comprising reasonably clean and natural English text.
The version we use, \textsc{C4.en}, consists of 365M English articles. Each comes with its text, URL, and crawling timestamp.
This amount of text enabled the pipeline to cover significantly more CSK assertions than the manually-built ConceptNet and any other automated CSK resource when limiting to top-100 assertions per subject (see Subsection~\ref{sec:assertion-recall}).
Our judicious rule-based approaches for extraction, filtering, and cleaning, as well as our unsupervised ranking, helped produce CSK assertions of higher saliency than any other automated CSK resource while maintaining a very high precision (see Subsection~\ref{sec:assertion-precision}).

\oldmaterial

We executed the new \ascentpp{} pipeline for all primary subjects in \ascent{} (i.e., 10K popular concept-centric subjects taken from ConceptNet). For each primary subject, we took its top 10 most frequent aspects and top 10 most frequent subgroups previously extracted by \ascent{} as high-quality fine-grained subjects. 

\subsection{Processing time and output size}
\label{sec:processing-time}
In Table~\ref{tab:statistics-pipeline}, we provide details on each step's processing time and outputs in the \ascentpp{} pipeline.

The corpus processing phase (Phase 1) is executed once, independent of any choice of subjects. We use SpaCy\footnote{\url{https://spacy.io/}}, a popular Python-based NLP library, as our NLP pipeline. First, we ran SpaCy on the 365M documents from C4 in parallel on a cluster of 6,400 CPU cores (AMD EPYC 7702 64-core processors), which took 1.5 days to complete. 
Then, it took 20h for the OpenIE system to digest all 356M processed documents on the same CPU cluster. The result of this step is a set of 8B OpenIE assertions.

Next, the filtering process took around 2 hours and resulted in 165M assertions for the selected subjects, which account for 15M unique triples. 

The clustering process, including precomputing embeddings and running the HAC algorithm, took around 10 hours. Embeddings were computed on a cluster of up to 150 GPUs (NVIDIA Quadro RTX 8000 GPUs) which took less than 2 hours. The HAC algorithm was run on the same CPU cluster, giving 7.5M clustered assertions. On average, two assertions form one cluster.

Next, the ConceptNet mapping module took around 30 minutes on the same GPU cluster. The rule-based cleaning step is the lightest component, taking less than 10 minutes on a personal laptop. It resulted in 2M assertions in the final \ascentpp{} CSKB, whose size is reported in Table~\ref{tab:statistics-subject-types}. In particular, there are about 10K subgroups and 5.8K aspects. We have 1.6M assertions for primary subjects, 80K assertions for subgroups, and 323K assertions for aspects. The total number of semantic facets is 2.3M. Hence, each assertion has more than one facet on average.  

Finally, the ranking module took less than 2 hours to complete on the same CPU and GPU clusters.

In summary, running the \ascentpp{} pipeline on the C4 corpus took approximately three days with our computing resources.

\addressreview
\subsection{Hyper-parameters}
\label{sec:hyperparameters}
For triple clustering, to compute the embeddings of our triples, we use the \textit{paraphrase-mpnet-base-v2}\footnote{\url{https://huggingface.co/sentence-transformers/paraphrase-mpnet-base-v2}} model, which was trained on multiple paraphrase datasets (e.g., \textit{altlex}, \textit{msmarco-triplets}, \textit{quora\_duplicates}, etc.) and has high performance in various sentence-embeddings evaluations\footnote{\url{https://sbert.net/docs/pretrained_models.html}}. 
The embedding vectors are normalized. For the HAC algorithm, we use Ward's linkage, and stop merging two clusters when their Euclidean distance exceeds 0.5. For facet clustering, the distance threshold is 1.0. These thresholds were chosen based on manual tuning on a small set of validation data points.

For the ConceptNet mapping, we fine-tuned the RoBERTa-base model~\cite{liu2019roberta} for 3 epochs using the AdamW optimizer with a batch size of 64 samples, a learning rate of $5e^{-5}$, and a weight decay of $0.001$.

\oldmaterial
For typicality scoring, we trained a regression model on 500 manually-annotated examples. The resulting formula is:
$\textit{typicality} = 0.324 \times \textit{modifier\_score} + 0.428 \times \textit{frequency} + 0.088 \times \textit{neutrality}$.

\begin{table}[t]
    \centering
    \begin{tabular}{lrrr}
    \toprule
    \textbf{Subject type} & \textbf{\#s} & \textbf{\#spo} & \textbf{\#facets} \\
    \midrule
    Primary & 8,067 & 1,651,455 & 1,975,385 \\
    Subgroup & 10,191 & 80,176 & 62,581 \\
    Aspect & 5,843 & 323,257 & 312,004 \\
    All & 24,101 & 2,054,888 & 2,349,970 \\
    \bottomrule
    \end{tabular}
    \caption{Size of \ascentpp{}.}
    \label{tab:statistics-subject-types}
\end{table}



\oldmaterial

\oldmaterial
\section{Evaluation}\label{sec:evaluation}

The evaluation of \ascentpp{} is centered on three research questions:
\begin{itemize}
    \item \textbf{RQ1:} Is the resulting CSKB of higher quality than existing resources?
    \item \textbf{RQ2:} Does (structured) CSK help in extrinsic use cases?
    \item \textbf{RQ3:} What are the quality and extrinsic value of facets?
\end{itemize}
We discuss each of these research questions in its own subsection.

\oldmaterial
\subsection{Intrinsic evaluation}
\label{sec:intrinsic}

To investigate RQ1, we instantiate \emph{quality} with the standard notions of precision and recall, splitting precision further up into the dimensions of \emph{typicality} and \emph{saliency}, measuring this way the degree of truth and the degree of relevance of assertions (see Section \ref{sec:knowledge-model} and also \cite{quasimodo}).


\paragraph{Evaluation metrics} 
We employ three evaluation metrics: 
\begin{enumerate}
    \item \textit{typicality,}
    \item \textit{saliency,}
    \item \textit{relative recall.}
\end{enumerate}
Knowledge base construction is typically evaluated by precision and recall. Following earlier work \cite{quasimodo}, we split \textit{precision} into two dimensions: \textit{typicality} (the degree of truth) and \textit{saliency} (how readily a statement is available to a human). These two dimensions are generally independent, as salient statements need not be typical, and vice versa. Furthermore, as the absolute recall is difficult to establish (e.g., there is no way to obtain a complete set of all assertions for what elephants are capable of doing), we use a \textit{relative recall} metric, measuring the fraction of statements from a human-built resource, that are captured in the respective CSKB.

\oldmaterial

\paragraph{Compared resources}
We compare \ascentpp{} with its predecessor \ascent{}~\cite{ascent}, as well as five other prominent resources:
\begin{enumerate}
    \item ConceptNet~\cite{conceptnet},
    \item TransOMCS~\cite{DBLP:conf/ijcai/ZhangKSR20},
    \item TupleKB~\cite{mishra2017domain},
    \item Quasimodo~\cite{quasimodo},
    \item COMET-ATOMIC$^{20}_{20}$~\cite{comet-atomic-2020} (with a caveat, see below).
\end{enumerate}
The ATOMIC~\cite{atomic}, AutoTOMIC~\cite{west2021symbolic}, and ASER~\cite{zhang2020aser} projects do not qualify for comparison as they do not contain concept-centric assertions.
ATOMIC$^{20}_{20}$~\cite{comet-atomic-2020} has a portion for physical commonsense, but most of those assertions come directly from ConceptNet (except for the \textit{ObjectUse} relation for which more human-annotated data was collected). Therefore, we do not include ATOMIC$^{20}_{20}$ directly in this evaluation. Instead, we compare other CSKBs with a generative model trained on that resource, the COMET-ATOMIC$^{20}_{20}$ model~\cite{comet-atomic-2020}.


\begin{table*}[t]
    \centering
    \begin{tabular}{llll}
    \toprule
    \textbf{Dimension} & \textbf{Question} & \textbf{Options} \\
    \midrule
    Typicality & \textit{Is this a correct assertion about \textless{}subject\textgreater{}?} & \begin{tabular}[c]{@{}l@{}}\textbf{Always/Often} - the knowledge assertion presented is always or often true\\ \textbf{Sometimes/Likely} - it is sometimes or likely true\\ \textbf{Farfetched/Never} - it is false or farfetched at best\\ \textbf{Invalid} - it is invalid or makes no sense \\ 
    \end{tabular} \\
    \midrule
    Saliency & \textit{\begin{tabular}[c]{@{}l@{}}Imagine you have 2 minutes time to explain\\ a kid about \textless{}subject\textgreater{}, would you mention\\ the following information?\end{tabular}} & \begin{tabular}[c]{@{}l@{}}\textbf{Absolutely} - the information is very interesting/important for that concept\\ \textbf{Probably} - it is quite good to know\\ \textbf{Maybe not} - it is not interesting or too obvious/uninteresting/boring\\ \textbf{Definitely not} - it is completely irrelevant/unimportant/wrong or makes no sense \\ 
    \end{tabular} \\
    \bottomrule
    \end{tabular}
    \caption{Crowdsourcing question templates for typicality and saliency evaluation of CSK assertions.}
    \label{tab:mturk-precision}
\end{table*}

\begin{table*}[t]
\centering
\begin{tabular}{lrrrrrrrrr}
\toprule
\multirow{4}{*}{\textbf{CSK resource}} & \multicolumn{2}{c}{\textbf{Precision (\%)}} & \multicolumn{6}{c}{\textbf{Relative recall (\%)}} & \textbf{Size}\\
\cmidrule(l){2-3} \cmidrule(l){4-9} \cmidrule(l){10-10}
& \multirow{2}{*}{\textbf{Saliency@10}} & \multirow{2}{*}{\textbf{Typicality (all)}} & \multicolumn{3}{c}{\textbf{Top-100 assertions/subject}} & \multicolumn{3}{c}{\textbf{All assertions}} &  \\
\cmidrule(l){4-6} \cmidrule(l){7-9} \cmidrule(l){10-10}
&  &  & $\tau = 0.96$ & $\tau = 0.98$ & $\tau = 1.00$ & $\tau = 0.96$ & $\tau = 0.98$ & $\tau = 1.00$ & \textbf{\#spo} \\
\bigtableline
\multicolumn{3}{l}{\textit{Crowdsourced}} & & & & & & & \\
\bigtableline
$~~~$ ConceptNet~\cite{conceptnet} & 79.2 & 96.0 & 5.29 & 3.39 & 1.10 & 5.47 & 3.53 & 1.13 & 0.5M \\
\bigtableline
\multicolumn{3}{l}{\textit{Generative}} & & & & & & &  \\
\bigtableline
$~~~$ COMET-ATOMIC$^{20}_{20}$~\cite{comet-atomic-2020} & 55.2 & 78.9 & - & - & - & - & - & - & - \\
\bigtableline
\multicolumn{3}{l}{\textit{Extractive}} & & & & & & & \\
\bigtableline
$~~~$ TransOMCS~\cite{DBLP:conf/ijcai/ZhangKSR20} & 40.4 & 51.4 & 4.04 & 3.24 & 1.85 & 19.70 & 16.47 & 9.06 & 18.5M \\
\bigtableline
$~~~$ TupleKB~\cite{mishra2017domain} & 36.0 & \textbf{92.0} & 3.57 & 1.99 & 0.41 & 4.32 & 2.49 & 0.58 & 0.3M \\
\bigtableline
$~~~$ Quasimodo~\cite{quasimodo} & 38.8 & 67.8 & 11.05 & 9.47 & 5.17 & \textbf{21.87} & \textbf{19.36} & \textbf{10.88} & 6.3M \\
\bigtableline
$~~~$ \ascent{}~\cite{ascent} & 60.0 & 79.2 & 9.60 & 7.02 & 3.17 & 17.09 & 12.62 & 5.82 & 8.6M \\
\bigtableline
$~~~$ \ascentpp{} & \textbf{68.8} & 88.4 & \textbf{12.70} & \textbf{10.70} & \textbf{5.90} & 17.54 & 14.64 & 7.95 & 2.0M \\
\bottomrule
\end{tabular}
\caption{Intrinsic evaluation results of different CSK resources.
}
\label{tab:eval-triple-quality}
\end{table*}

\subsubsection{Precision: typicality and saliency} 
\label{sec:assertion-precision}
\paragraph{Evaluation scheme}
Unlike the precision of encyclopedic knowledge (``The Lion King'' was either produced by Disney or not), the precision of CSK is generally not a binary concept, calling for more refined evaluation metrics. 
\newmaterial
Following the Quasimodo project~\cite{quasimodo}, we assessed the \textit{typicality} and \textit{saliency} of triples. Given a CSK triple, we asked annotators on Amazon MTurk to rate the triple along a 4-point Likert scale on each of the two dimensions.
We present the crowdsourcing questions and answer options in Table~\ref{tab:mturk-precision}. 
The crowdsourcing templates are inspired by those introduced in COMET-ATOMIC$^{20}_{20}$ \cite{comet-atomic-2020}.
We performed two separated MTurk tasks on the two dimensions. For each task, we randomly sampled 500 triples among 200 prominent subjects in four common domains: animal, occupation, engineering, and geography. For the saliency task, the sampling pool was limited to the top 10 assertions per subject.

Each MTurk task contained five CSK triples and was assigned to three different workers. Following \cite{comet-atomic-2020}, we also used human-readable language forms for triples in the fixed-schema CSKBs (i.e., ConceptNet, TransOMCS, and COMET). For \ascentpp{}, we used the open triples as the prompt display. 
Any triples that receive the first two labels for a dimension (see Table~\ref{tab:mturk-precision}) are ranked as \textit{typical/salient}.
The final judgment for a triple is based on a majority vote over the choices provided by the three annotators. 
Annotation quality was ensured by requiring the MTurk annotators to be Master workers with an all-time approval rate of over 90\%. The inter-rater agreement on the three labels measured by Fleiss' $\kappa$ is 0.33 (i.e., fair agreement~\cite{fleiss1973equivalence}).

\paragraph{Results}
We report the precision evaluation results in the left part of Table~\ref{tab:eval-triple-quality}. 

\addressreview
Among automatically-constructed CSKBs (i.e., TransOMCS~\cite{DBLP:conf/ijcai/ZhangKSR20}, TupleKB~\cite{mishra2017domain}, Quasimodo~\cite{quasimodo}, \ascent{}~\cite{ascent}, and \ascentpp{}), \ascentpp{} yields the most salient statements by a large margin (9 percentage points over its predecessor, and $>$13 percentage points over all others).

For the typicality, the new resource outperforms all but the domain-specific TupleKB ($>$10 percentage points over the others). TupleKB still wins by 4 percentage points, yet produces unsalient statements (-32 percentage points) and for the science domain only, based on high-quality textbooks, with no obvious way to scale beyond (see also recall evaluation, where \ascentpp{} has 4x more recall than TupleKB).
For example, while the top assertions for ``elephant'' in \ascentpp{} include \triple{elephant, \textit{IsA}, social animal} and \triple{elephant, \textit{HasProperty}, intelligent}, those in TupleKB include \triple{elephant, \textit{HasPart}, skin cell} and \triple{elephant, \textit{HasPart}, cell membrane}.
\oldmaterial

Furthermore, we also compared our KB with a generative CSKB construction method, COMET~\cite{bosselut2019comet}, specifically the COMET-ATOMIC$^{20}_{20}$ model~\cite{comet-atomic-2020}. 
COMET is an autoregressive language model fine-tuned on existing CSK resources and can be used to predict possible objects of given subject-predicate pairs.
Since COMET does not come with a materialized resource, we had to generate assertions ourselves. As there is no obvious stop criterion, we only evaluated the precision of top assertions but could not evaluate COMET's recall.
We used the provided BART model\footnote{\url{https://github.com/allenai/comet-atomic-2020/tree/master/models/comet_atomic2020_bart}}, which was trained on ATOMIC$^{20}_{20}$ (which also includes ConceptNet assertions). 
For each pair of subject and predicate, we asked COMET to predict top-5 objects.
We used the same sampling processes and MTurk templates described above for typicality and saliency evaluation. 
The evaluation results of COMET are also included in Table~\ref{tab:eval-triple-quality}.
\ascentpp{} clearly  performs better than 
the generative model,
even though both have seen comparable amounts of texts.
Some examples of the top assertions in \ascentpp{} and the ones generated by COMET-ATOMIC$^{20}_{20}$ are shown in Table~\ref{tab:compare-vs-comet}.
Although COMET has the flexibility of generating objects to any given subject-predicate pair, it still makes many incorrect predictions and produces notable redundancies. On the other hand, \ascentpp{}, which collected OpenIE assertions from several sources and aggregated them through various steps such as filtering, clustering, and cleaning, produces more correct and complementary assertions.

\begin{table}[t]
\centering
\begin{tabular}{lll}
    \toprule
    \textbf{Subject-Predicate} & \textbf{\ascentpp{}} & \textbf{COMET-ATOMIC$^{20}_{20}$} \\
    \midrule
    elephant - \textit{CapableOf} & \begin{tabular}[c]{@{}l@{}}- perform trick \thumsup{}\\ - eat grass \thumsup{}\\ - eat fruit \thumsup{}\\ - become agitated\\ - give ride \thumsup{}\end{tabular} & \begin{tabular}[c]{@{}l@{}}- climb tree\\ - walk on land \thumsup{}\\ - climb tree trunk\\ - walk on tree\\ - eat elephant\end{tabular} \\ 
    \midrule
    \begin{tabular}[c]{@{}l@{}}beer - \textit{Made(Up)Of}\\\end{tabular} & \begin{tabular}[c]{@{}l@{}}- hop \thumsup{}\\ - water \thumsup{}\\ - barley \thumsup{}\\ - yeast \thumsup{}\\ - grain \thumsup{}\end{tabular} & \begin{tabular}[c]{@{}l@{}}-  beer\\ -  alcohol \thumsup{}\\ -  drunk\\ -  drinking\\ -  drink\end{tabular} \\  
    \midrule
    \begin{tabular}[c]{@{}l@{}}laptop - \textit{UsedFor}/\\ \mbox{\ \ \ \ \ \ \ \ \ \ \ \ } \textit{ObjectUse}\end{tabular} & \begin{tabular}[c]{@{}l@{}}- work \thumsup{}\\ - gaming \thumsup{}\\ - office work \thumsup{}\\ - email \thumsup{}\\ - social media \thumsup{}\end{tabular} & \begin{tabular}[c]{@{}l@{}}- browse the internet \thumsup{}\\ -  use as a coaster\\ -  play games on \thumsup{}\\ -  use as a weapon\\ -  browse the web \thumsup{}\end{tabular} \\ 
    \bottomrule
\end{tabular}
\caption{Top-5 assertions of selected subject-predicate pairs in \ascentpp{} and COMET-ATOMIC$^{20}_{20}$~\cite{comet-atomic-2020}.} 
\label{tab:compare-vs-comet}
\end{table}

\subsubsection{Relative recall} 
\label{sec:assertion-recall}
\paragraph{Ground truth}
Evaluating recall requires a notion of ground truth. We use a \textit{relative recall} notion w.r.t. the statements contained in the CSLB property norm dataset~\cite{devereux2014centre}, which consists of short human-written sentences about salient properties of general concepts. There are 22.6K sentences expressing properties of 638 concepts in the dataset. The CSLB dataset could also be considered a CSK resource. However, due to its limited size, we did not include it in the comparisons with other CSKBs. Instead, we used it as ground truth for evaluating relative recall and the mask prediction task (see Section~\ref{subsec:extrinsic}). 

\paragraph{Assertion matching algorithm}
Since there are always different expressions of a CSK assertion in natural language, we allow soft matching between assertions in CSK resources and sentences in the ground-truth CSLB dataset. 
Specifically, for each CSLB sentence, we find the most similar assertion in the target CSK resource based on the cosine similarity between their embeddings (computed by SentenceTransformers~\cite{sbert}). 
That assertion will be considered a 
true positive w.r.t
the given CSLB fact 
only if their cosine similarity is greater than or equal to a predefined threshold $\tau$.
For example, if $\tau = 1.0$, only exact-match assertions are considered true positives.
When lowering $\tau$, we can match, e.g., ``elephant eats grass'' and ``elephant feeds on grasses''.

\paragraph{Results}
The relative recall evaluation results are shown in the right part of Table~\ref{tab:eval-triple-quality}, where we once show relative recall using the top-100 assertions per subject and once all. We report results at three similarity thresholds: $\tau=0.96$, $\tau=0.98$ and $\tau=1.0$.

\addressreview We find that \ascentpp{} yields the highest relative recall among all
automated 
resources of the same size (``top-100'' columns in Table \ref{tab:eval-triple-quality}), outperforming the next best KB, Quasimodo, by 14-15\% in relative recall. 
%
The gap to the manually built ConceptNet is even larger, where \ascentpp{} achieves 2 to 5 times higher relative recall, depending on the similarity threshold. 

Considering all statements from each resource, Quasimodo and TransOMCS appear to have a slight edge; yet this is only due to the precision-oriented thresholding of \ascentpp{} (2M assertions vs. 18.5M for TransOMCS and 6.3M for Quasimodo). 
Without
the cleaning phase
(see Section~\ref{subsec:cleaning}), 
the unfiltered \ascentpp{} variant would be the size-wise better point of comparison: with a size of 7.5M assertions, it achieves relative recall scores of 22.13\%, 18.02\%, and 9.37\% for the three thresholds $\tau \in \{0.96, 0.98, 1.0\}$, respectively.
We also sampled 500 triples from each of the three KBs (unfiltered-\ascentpp{}, Quasimodo and TransOMCS) for another typicality evaluation.
The obtained typicality scores are: 82.2\% for unfiltered-\ascentpp{}, 56.4\% for Quasimodo and 51.2\% for TransOMCS.
Hence, this variant of \ascentpp{} outperforms TransOMCS on both precision and recall, and it is 
better than
Quasimodo in 
precision 
while having 
similar
recall.
Moreover, in Subsection~\ref{sec:ascentpp-vs-ascent}, we will show that when increasing the size of \ascentpp{} to reach that of \ascent{}, we still reach competitive typicality scores.
This gives us the flexibility to tune for either precision or recall, by adjusting the cleaning phase.


These results confirm that large-scale extraction from web crawls can significantly outperform the recall of resources built from smaller, specifically selected document collections (\ascent{}, TupleKB) without sacrificing precision. Furthermore, the significant gains in precision over TransOMCS and Quasimodo come at a much lower decrease in recall.
\oldmaterial

\subsubsection{\ascentpp{} vs. \ascent{}}
\label{sec:ascentpp-vs-ascent}
%
\addressreview
To compare \ascentpp{} and its predecessor, \ascent{}, in addition to saliency@10 and typicality-all for the 200 common subjects, as well as the relative recall (see Table~\ref{tab:eval-triple-quality}), we perform three more evaluations: \textit{typicality@10},  \textit{typicality-random} and \textit{saliency-random} (see Table~\ref{tab:ascent-vs-ascentpp}).

For typicality@10, we perform the same sampling process as for saliency@10; the only difference is that \ascentpp{} assertions are now sorted by typicality score instead of saliency score. 

For typicality-random and saliency-random, 
a fair comparison must consider the different sizes of the two resources.
Thus, we randomly sample 500 subjects from both resources. For these 500 subjects, \ascent{} has 86,054 CSK assertions. For \ascentpp{} to have a similar number of assertions, we increase the limit for the maximal number of assertions per subject in \ascentpp{} (i.e., the last filter of the pipeline, see Section~\ref{subsec:cleaning}) to 7,250. That results in 86,065 CSK assertions in \ascentpp{} for the 500 random subjects. 
Now that the two resources have comparable sizes, we randomly pick up 500 triples from each resource and use them for the two crowd-sourcing evaluations (typicality-random and saliency-random). 
%

The results in Table~\ref{tab:ascent-vs-ascentpp} show 
that \ascentpp{} clearly outperforms the prior \ascent{} KB by a large margin regarding the typicality of both
top-ranked statements and randomly sampled ones.
Although the saliency scores of random samples drop significantly compared to those of the top-10 statements (see Table~\ref{tab:eval-triple-quality}), \ascentpp{} still outperforms \ascent{}. 
Combined with the results in Table~\ref{tab:eval-triple-quality}, the new \ascentpp{} KB consistently shows better quality than its predecessor \ascent{} KB, regarding both 
precision and relative recall.
\oldmaterial

\begin{table}[t]
\centering
\begin{tabular}{lrrr}
\toprule
\textbf{CSKB} & \textbf{Typicality@10} & \textbf{Typicality} & \textbf{Saliency} \\
\textbf{ } & \textbf{ } & \textbf{Random} & \textbf{Random} \\
\midrule
\ascent{}~\cite{ascent} & 87.6 & 65.6 & 40.6 \\
\midrule
\ascentpp{} & \textbf{93.6} & \textbf{82.0} & \textbf{44.6} \\
\bottomrule
\end{tabular}
\caption{Comparison of \ascent{} vs. \ascentpp{}.}
\label{tab:ascent-vs-ascentpp}
\end{table}

\subsubsection{Precision of ConceptNet mapping}
To evaluate the ConceptNet mapping module, we manually annotate 100 random samples from the final \ascentpp{} KB. For each sample, we mark it as a correct mapping if the fixed-schema triple preserves the meaning of the original triple. 
We obtained a precision of 96\%.

\oldmaterial
\subsection{Extrinsic evaluation}
\label{subsec:extrinsic}

To answer RQ2, we conduct a comprehensive evaluation of the contribution of commonsense knowledge to question answering (QA) via three different setups, all based on the idea of  priming pre-trained LMs with context~\cite{GPT3,guu2020realm,petroni2020context}:
\begin{enumerate}
    \item In \textit{masked prediction} (MP) \cite{petroni2019language}, we ask language models to predict single tokens in generic sentences.
    \item \newmaterial In \textit{autoregressive LM-based QA} (AR), we provide questions and let LMs generate arbitrary answer sentences. 
    \item \oldmaterial In \textit{span prediction} (SP), LMs select the best answers from provided CSKB content~\cite{lan2019albert}.
\end{enumerate}

\begin{table*}[t]
\centering
\begin{tabular}{cp{.53\textwidth}p{.35\textwidth}}
\toprule
\textbf{Setting} & \textbf{Input} & \textbf{Sample output} \\
\midrule
MP & Elephants eat {[}MASK{]}. {[}SEP{]} Elephants eat roots, grasses, fruit, and bark, and they eat a lot of these things. & everything (15.52\%), trees (15.32\%), plants (11.26\%) \\
\midrule
\multirow{4}{*}{AR} & Context: Elephants eat roots, grasses, fruit, and bark, and they eat a lot of these things. & Elephants eat a variety of different foods. \\
 & Question: What do elephants eat? & \\
 & Answer: & \\
\midrule
\multirow{3}{*}{SP} & \text{question}=``What do elephants eat?'' & \text{start}=14, \text{end}=46, \\
 & \text{context}=``Elephants eat roots, grasses, fruit, and bark, and they eat a lot of these things.'' & \text{answer}=``roots, grasses, fruit, and bark'' \\
\bottomrule
\end{tabular}
\caption{Examples of three QA settings (MP - masked prediction, AR - autoregressive LM-based QA, SP - span prediction). Sample output was given by RoBERTa (for MP), GPT-3 (for AR) and ALBERT (for SP).}
\label{tab:qa-settings}
\end{table*}

We illustrate all settings in Table~\ref{tab:qa-settings}. In all settings, LMs are provided with a context in the form of assertions taken from competitor CSKBs. 
These setups are motivated by the observation that priming language models with context can significantly influence their predictions~\cite{guu2020realm,petroni2020context}. Previous works on language model priming mainly focused on evaluating retrieval strategies. In contrast, our comprehensive test suite focuses on the impact of utilizing different CSK resources while leaving the retrieval component constant.

Masked prediction is perhaps the best-researched problem, coming with the advantage of allowing automated evaluation, although automated evaluation may unfairly discount sensible alternative answers. Also, masked prediction is limited to single tokens. 
Autoregressive LM-based generations circumvent this restriction, although they necessitate human annotations and can be prone to evasive answers. They are thus well complemented by extractive answering schemes, limiting the language models' abstraction abilities but providing the cleanest way to evaluate the context alone.

\paragraph{Models} Following standard usage, we use RoBERTa-large~\cite{liu2019roberta} for masked prediction, GPT-3~\cite{GPT3} for the generative setup, and ALBERT-xxlarge~\cite{lan2019albert}, fine-tuned on SQuAD 2.0, for span prediction.

\newmaterial
\paragraph{Context retrieval method}
Given a query, we use sentence embeddings to pull out relevant assertions from a CSKB.
Using the SentenceTransformers model \textit{msmarco-distilbert-base-v3}\footnote{\url{https://www.sbert.net/docs/pretrained-models/msmarco-v3.html}}, we compute embeddings of the given query and all lexicalized triples in the CSKBs. Then we use cosine similarity to select the top 5 most similar triples to the query as context.

\oldmaterial
\paragraph{Task construction}
Previous work has generated masked sentences based on templates from ConceptNet triples~\cite{petroni2019language}. However, the resulting sentences are often unnatural, following the idiosyncrasies of the ConceptNet data model. 
\newmaterial
Therefore, we created a new dataset of natural commonsense sentences for \textit{masked prediction} that is based on an independent human-annotated dataset, the aforementioned CSLB dataset~\cite{devereux2014centre}, and consists of 19.6K masked sentences~\cite{ascent}.

\oldmaterial
For the \textit{generative} and \textit{extractive} settings, we used the Google Search auto-completion functionality to collect commonsense questions about common subjects by feeding the API with six prefixes: ``what/when/where are/do $<$subject$>$...''.
That process returned 8,098 auto-completed queries.
\newmaterial
Next, we drew samples from the query set, then manually removed jokes and other noise (e.g., ``where do cows go for entertainment''), obtaining 100 questions for evaluation.

\oldmaterial
\paragraph{Evaluation scheme}
For commonsense topics, questions often have multiple valid answers. Additionally, given that answers in our generative and extractive QA settings are very open, creating an automated evaluation is difficult. We, therefore, use human judgments for evaluating all settings except masked prediction. Specifically, given a question and set of answers, we ask human annotators to assess each answer based on two dimensions, \textit{correctness}, and \textit{informativeness}, 
each on a 4-point Likert scale from 0 (lowest) to 3 (highest) (see Table~\ref{tab:mturk-qa} for the question templates given to the annotators). Correctness indicates whether an answer holds true. Informativeness reflects that the information conveyed by an answer is helpful.

An answer could be correct and uninformative at the same time. For example, given the question ``What do elephants use their trunk for?'', the answers ``For breathing.'' and ``To suck up water.'' are both correct and informative. However, ``To do things.'' is a correct answer but not informative at all. On the other hand, if an answer is incorrect, it should automatically be uninformative.

\begin{table*}[t]
    \centering
    \begin{tabular}{llll}
    \toprule
    \textbf{Dimension} & \textbf{Question} & \textbf{Options} \\
    \midrule
    Correctness & \textit{How often does this answer hold true?} & \begin{tabular}[c]{@{}l@{}}\textbf{3 - Always/Often} - the answer is always or often true\\ \textbf{2 - Sometimes/Likely} - it is sometimes or likely true\\ \textbf{1 - Farfetched/Never} - it is false or farfetched at best\\ \textbf{0 - Invalid} - it is invalid or makes no sense\end{tabular} \\
    \midrule
    Informativeness & \textit{How informative is the answer?} & \begin{tabular}[c]{@{}l@{}}\textbf{3 - Highly} - the answer provides very useful knowledge w.r.t the question\\ \textbf{2 - Moderately} - the answer is moderately useful\\ \textbf{1 - Slightly} - the answer is slightly useful\\ \textbf{0 - Not at all} - the answer is too general or makes no sense\end{tabular} \\
    \bottomrule
    \end{tabular}
    \caption{Crowdsourcing question templates for evaluation of the AR and SP settings for commonsense QA.}
    \label{tab:mturk-qa}
\end{table*}

Three annotators evaluate each question in Amazon MTurk with the same qualification requirements as in Section~\ref{sec:intrinsic}. We use the mean precision at $k$ ($P@k$) metric for evaluating masked prediction, following \cite{petroni2019language}.

\newmaterial

\paragraph{Results}
The evaluation results are shown in Table~\ref{tab:eval-qa}.

For \textit{masked prediction}, all CSKBs contribute useful contexts that substantially improve the quality of LM responses. \ascentpp{}, along with the only manually-constructed KB, ConceptNet, statistically significantly outperforms all other KBs at every threshold $k$ (all p-values of paired t-test below 0.05).

For \textit{autoregressive LM-based QA}, 
markedly, we find that GPT-3 performs on average better without any context. However, in combination with \ascentpp{}, it still performs better than with any other CSKB. More research is needed to design methods to pull relevant context from CSKBs and decide when to use it in LM-based QA and when to rely on the LM's knowledge alone.


For \textit{span prediction}, where answers come directly from retrieved contexts, \ascentpp{} also outperforms all other competitors. \ascentpp{} obtained statistically significant gains over Quasimodo, TupleKB, and TransOMCS on both metrics, and over ConceptNet on correctness. This indicates that our \ascentpp{} assertions have high quality compared to others.

\begin{table}[t]
\centering
\begin{tabular}{lrrrrrrr}
\toprule
\multicolumn{1}{l}{\multirow{2}{*}{\textbf{Context}}} & \multicolumn{3}{c}{\textbf{MP}} & \multicolumn{2}{c}{\textbf{AR}} & \multicolumn{2}{c}{\textbf{SP}} \\
\cmidrule{2-8}
\multicolumn{1}{c}{} & \textbf{P@1} & \textbf{P@5} & \textbf{P@10} & \textbf{C} & \textbf{I} & \textbf{C} & \textbf{I} \\
\midrule
No context & 8.10 & 17.16 & 21.37 & \textbf{2.47} & \textbf{2.01} & - & - \\
\midrule
ConceptNet & \textbf{14.41} & \textbf{27.08} & 32.16 & 2.22 & 1.70 & 1.74 & 1.52 \\
\midrule
TransOMCS & 7.08 & 15.42 & 19.99 & 1.32 & 0.86 & 0.99 & 0.85 \\
\midrule
TupleKB & 11.61 & 24.76 & 30.36 & 2.22 & 1.51 & 1.70 & 1.38 \\
\midrule
Quasimodo & 12.11 & 22.75 & 27.71 & 2.03 & 1.51 & 1.75 & 1.44 \\
\midrule
\ascent{} & 11.95 & 24.70 & 29.70 & 2.25 & 1.76 & 1.88 & 1.60 \\
\midrule
\ascentpp{} & 13.30 & 27.03 & \textbf{32.90} & 2.32 & 1.71 & \textbf{1.94} & \textbf{1.63} \\
\bottomrule
\end{tabular}
\caption{QA evaluation results. Metrics: \textbf{P@k} - mean precision at $k$ (\%), \textbf{C} - correctness ([0, 3]), \textbf{I} - informativeness ([0, 3]). 
}
\label{tab:eval-qa}
\end{table}

\subsection{Evaluation of facets}


To answer RQ3, we evaluate facets both intrinsically and extrinsically. 

\paragraph{Intrinsic evaluation}
As there are no existing CSKBs coming with facets, we provide comparisons with 
a
strong LM 
baseline,
GPT-2~\cite{radford2019language}.
First, we randomly drew 300 assertions along with their top-1 facets from our KB. 
Next, we translate each statement into a sentence prefix and ask 
GPT-2
to fill in the remaining words to complete the sentence. For example, given the quadruple \triple{elephant, use, their trunks, \textsc{purpose}: to suck up water}, the sentence prefix will be ``Elephants use their trunk to'' and for this, GPT-2's continuation is ``to move around.''
Then, each sentence prefix along with 
two
answers (from \ascentpp{}
and
GPT-2)
were shown to a human annotator (without knowing the source of the answers) who
annotated if each answer was
\textit{correct/incorrect}
and
\textit{informative/uninformative},
following similar metrics used for the QA evaluation in Section~\ref{subsec:extrinsic}.

The results are reported in Table~\ref{tab:facet-assessment}. 
\ascentpp{} achieves  70.10\% correctness and 54.15\% informativeness, both significantly better than the values for the GPT-2 model. 

\begin{table}[t]
    \centering
    \small
    \begin{tabular}{rrr}
        \toprule
         & \textbf{Correct (\%)} & \textbf{Informative (\%)} \\
        \midrule
        GPT-2~\cite{radford2019language} & 61.79 & 48.84 \\
        \ascentpp{} & \textbf{70.10} & \textbf{54.15} \\
        \bottomrule
    \end{tabular}
    \caption{Assessment of \ascentpp{} facets and LM-generated facets.} 
    \label{tab:facet-assessment}
\end{table}

\paragraph{Extrinsic evaluation}
We reused the three question answering tasks from Section~\ref{subsec:extrinsic}. 
The results are shown in Table~\ref{tab:facet-extrinsic}.
Expanding the \ascentpp{} triples with facets gives a consistent improvement in four of five evaluation metrics (precision at one in MP, informativeness in AR and SP, and correctness in SP), with the most prominent effect being observed for span prediction (8.6\% and 9.3\% relative improvements over the no-facet context in informativeness and correctness, respectively).

\begin{table}[t]
\centering
\begin{tabular}{lrrrrr}
\toprule
\multicolumn{1}{l}{\multirow{2}{*}{\textbf{Context}}} & \textbf{MP} & \multicolumn{2}{c}{\textbf{AR}} & \multicolumn{2}{c}{\textbf{SP}} \\
\cmidrule{2-6}
\multicolumn{1}{c}{} & \textbf{P@1} & \textbf{C} & \textbf{I} & \textbf{C} & \textbf{I} \\
\midrule
Without facets & 13.30 & \textbf{2.32} & 1.71 & 1.94 & 1.63 \\
\midrule
With facets & \textbf{13.38} & 2.26 & \textbf{1.79} & \textbf{2.12} & \textbf{1.77} \\
\bottomrule
\end{tabular}
\caption{Extrinsic evaluation of facets by mean precision at one (\%), correctness \textbf{C} ([0-3]) and informativeness \textbf{I} ([0-3]).}
\label{tab:facet-extrinsic}
\end{table}



\section{Conclusion}

This paper presented \ascentpp{}, a methodology to 
extract and semantically organize
advanced commonsense knowledge from large-scale web contents.
Our refined knowledge representation allowed us to identify considerably more informative assertions, overcoming the limitations of prior works. The
techniques for filtering, aggregating, and consolidating extracted tuples show that CSK extraction from broad web content is feasible at scale, with both high precision and high recall. Intrinsic and extrinsic evaluations confirmed that the resulting CSKB is a significant advance over existing CSK collections and provides an edge over recent language-model-based approaches. 
Code, data, and a web interface are 
accessible
at \textit{\url{\websiteurl}}.

\oldmaterial

\bibliographystyle{IEEEtran}
\bibliography{refs}

\begin{thebibliography}{10}
\providecommand{\url}[1]{#1}
\csname url@samestyle\endcsname
\providecommand{\newblock}{\relax}
\providecommand{\bibinfo}[2]{#2}
\providecommand{\BIBentrySTDinterwordspacing}{\spaceskip=0pt\relax}
\providecommand{\BIBentryALTinterwordstretchfactor}{4}
\providecommand{\BIBentryALTinterwordspacing}{\spaceskip=\fontdimen2\font plus
\BIBentryALTinterwordstretchfactor\fontdimen3\font minus
  \fontdimen4\font\relax}
\providecommand{\BIBforeignlanguage}[2]{{%
\expandafter\ifx\csname l@#1\endcsname\relax
\typeout{** WARNING: IEEEtran.bst: No hyphenation pattern has been}%
\typeout{** loaded for the language `#1'. Using the pattern for}%
\typeout{** the default language instead.}%
\else
\language=\csname l@#1\endcsname
\fi
#2}}
\providecommand{\BIBdecl}{\relax}
\BIBdecl

\bibitem{mccarthy1960programs}
J.~McCarthy, \emph{Programs with common sense}.\hskip 1em plus 0.5em minus
  0.4em\relax RLE and MIT computation center, 1960.

\bibitem{feigenbaum1984knowledge}
E.~A. Feigenbaum, ``Knowledge engineering,'' \emph{Annals of the New York
  Academy of Sciences}, 1984.

\bibitem{lenat1995cyc}
D.~B. Lenat, ``Cyc: A large-scale investment in knowledge infrastructure,''
  \emph{{CACM}}, 1995.

\bibitem{OpenCyc}
``{The OpenCyc Platform},''
  \url{https://web.archive.org/web/20160330064354/http://www.opencyc.org/},
  accessed: 2022-05-16.

\bibitem{conceptnet}
R.~Speer and C.~Havasi, ``{ConceptNet} 5: A large semantic network for
  relational knowledge,'' \emph{Theory and Applications of Natural Language
  Processing}, 2012.

\bibitem{webchild}
N.~Tandon, G.~de~Melo, F.~M. Suchanek, and G.~Weikum, ``{WebChild}: harvesting
  and organizing commonsense knowledge from the web,'' in \emph{WSDM}, 2014.

\bibitem{mishra2017domain}
B.~D. Mishra, N.~Tandon, and P.~Clark, ``Domain-targeted, high precision
  knowledge extraction,'' \emph{TACL}, 2017.

\bibitem{quasimodo}
J.~Romero, S.~Razniewski, K.~Pal, J.~Z. Pan, A.~Sakhadeo, and G.~Weikum,
  ``Commonsense properties from query logs and question answering forums,'' in
  \emph{CIKM}, 2019.

\bibitem{Panton2006CommonSR}
K.~Panton, C.~Matuszek, D.~B. Lenat, D.~Schneider, M.~Witbrock, N.~Siegel, and
  B.~Shepard, ``Common sense reasoning - from cyc to intelligent assistant,''
  in \emph{Ambient Intelligence in Everyday}, 2006.

\bibitem{lin2017reasoning}
H.~Lin, L.~Sun, and X.~Han, ``Reasoning with heterogeneous knowledge for
  commonsense machine comprehension,'' in \emph{EMNLP}, 2017.

\bibitem{lin2019kagnet}
B.~Y. Lin, X.~Chen, J.~Chen, and X.~Ren, ``Kagnet: Knowledge-aware graph
  networks for commonsense reasoning,'' in \emph{EMNLP-IJCNLP}, 2019.

\bibitem{xia2019incorporating}
J.~Xia, C.~Wu, and M.~Yan, ``Incorporating relation knowledge into commonsense
  reading comprehension with multi-task learning,'' in \emph{CIKM}, 2019.

\bibitem{lin-etal-2020-commongen}
B.~Y. Lin, W.~Zhou, M.~Shen, P.~Zhou, C.~Bhagavatula, Y.~Choi, and X.~Ren,
  ``{C}ommon{G}en: A constrained text generation challenge for generative
  commonsense reasoning,'' in \emph{Findings of EMNLP}, 2020.

\bibitem{ilievski2020commonsense}
F.~Ilievski, P.~Szekely, and D.~Schwabe, ``Commonsense knowledge in
  {Wikidata},'' in \emph{Wikidata workshop}, 2020.

\bibitem{atomic}
M.~Sap, R.~LeBras, E.~Allaway, C.~Bhagavatula, N.~Lourie, H.~Rashkin, B.~Roof,
  N.~A. Smith, and Y.~Choi, ``Atomic: An atlas of machine commonsense for
  if-then reasoning,'' in \emph{AAAI}, 2018.

\bibitem{DBLP:conf/ijcai/ZhangKSR20}
H.~Zhang, D.~Khashabi, Y.~Song, and D.~Roth, ``Transomcs: From linguistic
  graphs to commonsense knowledge,'' in \emph{IJCAI}, 2020.

\bibitem{west2021symbolic}
P.~West, C.~Bhagavatula, J.~Hessel, J.~D. Hwang, L.~Jiang, R.~L. Bras, X.~Lu,
  S.~Welleck, and Y.~Choi, ``Symbolic knowledge distillation: from general
  language models to commonsense models,'' in \emph{NAACL}, 2022.

\bibitem{t5}
C.~Raffel, N.~Shazeer, A.~Roberts, K.~Lee, S.~Narang, M.~Matena, Y.~Zhou,
  W.~Li, and P.~J. Liu, ``Exploring the limits of transfer learning with a
  unified text-to-text transformer,'' \emph{Journal of Machine Learning
  Research}, 2020.

\bibitem{ascent}
T.-P. Nguyen, S.~Razniewski, and G.~Weikum, ``Advanced semantics for
  commonsense knowledge extraction,'' in \emph{WWW}, 2021.

\bibitem{sbert}
N.~Reimers and I.~Gurevych, ``Sentence-bert: Sentence embeddings using siamese
  bert-networks,'' in \emph{EMNLP}, 2019.

\bibitem{GPT3}
T.~B. Brown \emph{et~al.}, ``Language models are few-shot learners,'' in
  \emph{NeurIPS}, 2020.

\bibitem{singh2002open}
P.~Singh, T.~Lin, E.~T. Mueller, G.~Lim, T.~Perkins, and W.~L. Zhu, ``Open mind
  common sense: Knowledge acquisition from the general public,'' in \emph{OTM
  Confederated International Conferences}, 2002.

\bibitem{liu2004conceptnet}
H.~Liu and P.~Singh, ``Conceptnet—a practical commonsense reasoning
  tool-kit,'' \emph{BT technology journal}, 2004.

\bibitem{DBLP:conf/aaai/GordonDS10}
J.~Gordon, B.~V. Durme, and L.~K. Schubert, ``Learning from the web: Extracting
  general world knowledge from noisy text,'' in \emph{{AAAI} Workshops}, 2010.

\bibitem{comet-atomic-2020}
J.~D. Hwang, C.~Bhagavatula, R.~L. Bras, J.~Da, K.~Sakaguchi, A.~Bosselut, and
  Y.~Choi, ``{(Comet-)Atomic} 2020: On symbolic and neural commonsense
  knowledge graphs,'' in \emph{AAAI}, 2021.

\bibitem{zhang2020aser}
H.~Zhang, X.~Liu, H.~Pan, Y.~Song, and C.~W.-K. Leung, ``Aser: A large-scale
  eventuality knowledge graph,'' in \emph{WWW}, 2020.

\bibitem{aser2}
H.~Zhang, X.~Liu, H.~Pan, H.~Ke, J.~Ou, T.~Fang, and Y.~Song, ``{ASER}: Towards
  large-scale commonsense knowledge acquisition via higher-order selectional
  preference over eventualities,'' \emph{Artificial Intelligence}, 2022.

\bibitem{ilievski2021cskg}
F.~Ilievski, P.~Szekely, and B.~Zhang, ``Cskg: The commonsense knowledge
  graph,'' in \emph{ESWC}, 2021.

\bibitem{bhakthavatsalam2020genericskb}
S.~Bhakthavatsalam, C.~Anastasiades, and P.~Clark, ``Genericskb: A knowledge
  base of generic statements,'' \emph{arXiv:2005.00660}, 2020.

\bibitem{ascent-demo}
T.-P. Nguyen, S.~Razniewski, and G.~Weikum, ``Inside {ASCENT}: Exploring a deep
  commonsense knowledge base and its usage in question answering,'' in
  \emph{ACL: System Demonstrations}, 2021.

\bibitem{DBLP:conf/www/EtzioniCDKPSSWY04}
O.~Etzioni, M.~J. Cafarella, D.~Downey, S.~Kok, A.~Popescu, T.~Shaked,
  S.~Soderland, D.~S. Weld, and A.~Yates, ``Web-scale information extraction in
  knowitall: (preliminary results),'' in \emph{WWW}, 2004.

\bibitem{DBLP:conf/acl/SnowJN06}
R.~Snow, D.~Jurafsky, and A.~Y. Ng, ``Semantic taxonomy induction from
  heterogenous evidence,'' in \emph{ACL}, 2006.

\bibitem{DBLP:journals/coling/GirjuBM06}
R.~Girju, A.~Badulescu, and D.~I. Moldovan, ``Automatic discovery of part-whole
  relations,'' \emph{Comput. Linguistics}, 2006.

\bibitem{DBLP:conf/acl/PantelP06}
P.~Pantel and M.~Pennacchiotti, ``Espresso: Leveraging generic patterns for
  automatically harvesting semantic relations,'' in \emph{ACL}, 2006.

\bibitem{DBLP:conf/acl/PascaD08}
M.~Pasca and B.~V. Durme, ``Weakly-supervised acquisition of open-domain
  classes and class attributes from web documents and query logs,'' in
  \emph{ACL}, 2008.

\bibitem{DBLP:journals/ai/PonzettoS11}
S.~P. Ponzetto and M.~Strube, ``Taxonomy induction based on a collaboratively
  built knowledge repository,'' \emph{Artif. Intell.}, 2011.

\bibitem{DBLP:conf/sigmod/WuLWZ12}
W.~Wu, H.~Li, H.~Wang, and K.~Q. Zhu, ``Probase: a probabilistic taxonomy for
  text understanding,'' in \emph{SIGMOD}, 2012.

\bibitem{HertlingPaulheim:ISWC2017}
S.~Hertling and H.~Paulheim, ``{WebIsALOD}: Providing hypernymy relations
  extracted from the web as linked open data,'' in \emph{ISWC}, 2017.

\bibitem{wordnet}
G.~A. Miller, ``Wordnet: A lexical database for {English},'' \emph{CACM}, 1995.

\bibitem{seitner2016large}
J.~Seitner, C.~Bizer, K.~Eckert, S.~Faralli, R.~Meusel, H.~Paulheim, and S.~P.
  Ponzetto, ``A large database of hypernymy relations extracted from the web,''
  in \emph{LREC}, 2016.

\bibitem{DBLP:conf/sigmod/LiuGNLWWX20}
B.~Liu, W.~Guo, D.~Niu, J.~Luo, C.~Wang, Z.~Wen, and Y.~Xu, ``{GIANT:} scalable
  creation of a web-scale ontology,'' in \emph{SIGMOD}, 2020.

\bibitem{tandon2016commonsense}
N.~Tandon, C.~Hariman, J.~Urbani, A.~Rohrbach, M.~Rohrbach, and G.~Weikum,
  ``Commonsense in parts: Mining part-whole relations from the web and image
  tags,'' in \emph{AAAI}, 2016.

\bibitem{haspartkb}
S.~Bhakthavatsalam, K.~Richardson, N.~Tandon, and P.~Clark, ``Do dogs have
  whiskers? a new knowledge base of haspart relations,''
  \emph{arXiv:2006.07510}, 2020.

\bibitem{gabbay2003many}
D.~M. Gabbay, \emph{Many-Dimensional Modal Logics: Theory and
  Applications}.\hskip 1em plus 0.5em minus 0.4em\relax Elsevier North Holland,
  2003.

\bibitem{hoganetal}
A.~Hogan \emph{et~al.}, ``Knowledge graphs,'' \emph{ACM Computing Surveys
  (CSUR)}, 2021.

\bibitem{schubert2002can}
L.~Schubert, ``Can we derive general world knowledge from texts,'' in
  \emph{HLT}, 2002.

\bibitem{DBLP:journals/expert/DragoniPC18}
M.~Dragoni, S.~Poria, and E.~Cambria, ``Ontosenticnet: {A} commonsense ontology
  for sentiment analysis,'' \emph{{IEEE} Intell. Syst.}, 2018.

\bibitem{zhang2017ordinal}
S.~Zhang, R.~Rudinger, K.~Duh, and B.~Van~Durme, ``Ordinal common-sense
  inference,'' \emph{TACL}, 2017.

\bibitem{chen-etal-2020-uncertain}
T.~Chen, Z.~Jiang, A.~Poliak, K.~Sakaguchi, and B.~Van~Durme, ``Uncertain
  natural language inference,'' in \emph{ACL}, 2020.

\bibitem{chalier2020joint}
Y.~Chalier, S.~Razniewski, and G.~Weikum, ``Joint reasoning for multi-faceted
  commonsense knowledge,'' in \emph{AKBC}, 2020.

\bibitem{DBLP:series/synthesis/2010Palmer}
M.~Palmer, D.~Gildea, and N.~Xue, \emph{Semantic Role Labeling}, ser. Synthesis
  Lectures on Human Language Technologies.\hskip 1em plus 0.5em minus
  0.4em\relax Morgan {\&} Claypool Publishers, 2010.

\bibitem{clarke2012nlp}
J.~Clarke, V.~Srikumar, M.~Sammons, and D.~Roth, ``An {NLP} curator (or: How i
  learned to stop worrying and love {NLP} pipelines),'' in \emph{LREC}, 2012.

\bibitem{semanticrolelabelling}
G.~Stanovsky, J.~Michael, L.~Zettlemoyer, and I.~Dagan, ``Supervised open
  information extraction,'' in \emph{NAACL}, 2018.

\bibitem{stuffie}
R.~E. Prasojo, M.~Kacimi, and W.~Nutt, ``Stuffie: Semantic tagging of unlabeled
  facets using fine-grained information extraction,'' in \emph{CIKM}, 2018.

\bibitem{graphene}
M.~Cetto, C.~Niklaus, A.~Freitas, and S.~Handschuh, ``Graphene:
  Semantically-linked propositions in open information extraction,'' in
  \emph{COLING}, 2018.

\bibitem{devlin2019bert}
J.~Devlin, M.-W. Chang, K.~Lee, and K.~Toutanova, ``Bert: Pre-training of deep
  bidirectional transformers for language understanding,'' in \emph{NAACL},
  2019.

\bibitem{radford2019language}
A.~Radford, J.~Wu, R.~Child, D.~Luan, D.~Amodei, I.~Sutskever \emph{et~al.},
  ``Language models are unsupervised multitask learners,'' \emph{OpenAI blog},
  vol.~1, no.~8, p.~9, 2019.

\bibitem{petroni2019language}
F.~Petroni, T.~Rockt{\"a}schel, P.~Lewis, A.~Bakhtin, Y.~Wu, A.~H. Miller, and
  S.~Riedel, ``Language models as knowledge bases?'' in \emph{EMNLP}, 2019.

\bibitem{bosselut2019comet}
A.~Bosselut, H.~Rashkin, M.~Sap, C.~Malaviya, A.~{\c{C}}elikyilmaz, and
  Y.~Choi, ``{COMET:} commonsense transformers for automatic knowledge graph
  construction,'' in \emph{ACL}, 2019.

\bibitem{cometalt1}
J.~Davison, J.~Feldman, and A.~M. Rush, ``Commonsense knowledge mining from
  pretrained models,'' in \emph{EMNLP}, 2019.

\bibitem{cometalt2}
Z.~Bouraoui, J.~Camacho-Collados, and S.~Schockaert, ``Inducing relational
  knowledge from bert,'' in \emph{AAAI}, 2020.

\bibitem{nguyen-razniewski-2022-materialized}
T.-P. Nguyen and S.~Razniewski, ``Materialized knowledge bases from commonsense
  transformers,'' in \emph{Proceedings of the First Workshop on Commonsense
  Representation and Reasoning}, 2022.

\bibitem{DBLP:journals/csur/Navigli09}
R.~Navigli, ``Word sense disambiguation: {A} survey,'' \emph{{ACM} Comput.
  Surv.}, 2009.

\bibitem{DBLP:conf/aaaifs/GordonS10}
J.~Gordon and L.~K. Schubert, ``Quantificational sharpening of commonsense
  knowledge,'' in \emph{Commonsense Knowledge, {AAAI} Fall Symposium}, 2010.

\bibitem{shwartz2018olive}
V.~Shwartz and C.~Waterson, ``Olive oil is made of olives, baby oil is made for
  babies: Interpreting noun compounds using paraphrases in a neural model,'' in
  \emph{NAACL}, 2018.

\bibitem{clark2018think}
P.~Clark, I.~Cowhey, O.~Etzioni, T.~Khot, A.~Sabharwal, C.~Schoenick, and
  O.~Tafjord, ``Think you have solved question answering? try {ARC}, the {AI2}
  reasoning challenge,'' \emph{arXiv:1803.05457}, 2018.

\bibitem{DBLP:conf/cikm/GordonD13}
J.~Gordon and B.~V. Durme, ``Reporting bias and knowledge acquisition,'' in
  \emph{AKBC}, 2013.

\bibitem{quasimodo-demo}
J.~Romero and S.~Razniewski, ``Inside quasimodo: Exploring construction and
  usage of commonsense knowledge,'' in \emph{CIKM}, 2020.

\bibitem{elazar2019large}
Y.~Elazar, A.~Mahabal, D.~Ramachandran, T.~Bedrax-Weiss, and D.~Roth, ``How
  large are lions? inducing distributions over quantitative attributes,'' in
  \emph{ACL}, 2019.

\bibitem{feng2020scalable}
Y.~Feng, X.~Chen, B.~Y. Lin, P.~Wang, J.~Yan, and X.~Ren, ``Scalable multi-hop
  relational reasoning for knowledge-aware question answering,'' in
  \emph{EMNLP}, 2020.

\bibitem{mikolov2013distributed}
T.~Mikolov, I.~Sutskever, K.~Chen, G.~S. Corrado, and J.~Dean, ``Distributed
  representations of words and phrases and their compositionality,'' in
  \emph{NeurIPS}, 2013.

\bibitem{wang-etal-2021-phrase}
S.~Wang, L.~Thompsondo, and M.~Iyyer, ``Phrase-{BERT}: Improved phrase
  embeddings from {BERT} with an application to corpus exploration,'' in
  \emph{EMNLP}, 2021.

\bibitem{yago4}
T.~Pellissier~Tanon, G.~Weikum, and F.~Suchanek, ``Yago 4: A reason-able
  knowledge base,'' in \emph{ESWC}, 2020.

\bibitem{soderland2013open}
S.~Soderland, J.~Gilmer, R.~Bart, O.~Etzioni, and D.~S. Weld, ``Open
  information extraction to kbp relations in 3 hours,'' in \emph{TAC}, 2013.

\bibitem{galarraga2014canonicalizing}
L.~Gal{\'a}rraga, G.~Heitz, K.~Murphy, and F.~M. Suchanek, ``Canonicalizing
  open knowledge bases,'' in \emph{CIKM}, 2014.

\bibitem{putri2019aligning}
R.~A. Putri, G.~Hong, and S.-H. Myaeng, ``Aligning open ie relations and kb
  relations using a siamese network based on word embedding,'' in \emph{IWCS},
  2019.

\bibitem{zhang2019openki}
D.~Zhang, S.~Mukherjee, C.~Lockard, X.~L. Dong, and A.~McCallum, ``Openki:
  Integrating open information extraction and knowledge bases with relation
  inference,'' in \emph{NAACL}, 2019.

\bibitem{navigli2012babelnet}
R.~Navigli and S.~P. Ponzetto, ``Babelnet: The automatic construction,
  evaluation and application of a wide-coverage multilingual semantic
  network,'' \emph{Artificial intelligence}, 2012.

\bibitem{mehrabi2021lawyers}
N.~Mehrabi, P.~Zhou, F.~Morstatter, J.~Pujara, X.~Ren, and A.~Galstyan,
  ``Lawyers are dishonest? quantifying representational harms in commonsense
  knowledge resources,'' in \emph{EMNLP}, 2021.

\bibitem{liu2019roberta}
Y.~Liu, M.~Ott, N.~Goyal, J.~Du, M.~Joshi, D.~Chen, O.~Levy, M.~Lewis,
  L.~Zettlemoyer, and V.~Stoyanov, ``Roberta: A robustly optimized bert
  pretraining approach,'' \emph{arXiv:1907.11692}, 2019.

\bibitem{fleiss1973equivalence}
J.~L. Fleiss and J.~Cohen, ``The equivalence of weighted kappa and the
  intraclass correlation coefficient as measures of reliability,''
  \emph{Educational and psychological measurement}, 1973.

\bibitem{devereux2014centre}
B.~J. Devereux, L.~K. Tyler, J.~Geertzen, and B.~Randall, ``The centre for
  speech, language and the brain ({CSLB}) concept property norms,''
  \emph{Behavior research methods}, 2014.

\bibitem{guu2020realm}
K.~Guu, K.~Lee, Z.~Tung, P.~Pasupat, and M.-W. Chang, ``Realm:
  Retrieval-augmented language model pre-training,'' in \emph{ICML}, 2020.

\bibitem{petroni2020context}
F.~Petroni, P.~Lewis, A.~Piktus, T.~Rockt{\"a}schel, Y.~Wu, A.~H. Miller, and
  S.~Riedel, ``How context affects language models' factual predictions,'' in
  \emph{AKBC}, 2020.

\bibitem{lan2019albert}
Z.~Lan, M.~Chen, S.~Goodman, K.~Gimpel, P.~Sharma, and R.~Soricut, ``Albert: A
  lite {Bert} for self-supervised learning of language representations,'' in
  \emph{ICLR}, 2020.

\end{thebibliography}



\end{document}